\documentclass[journal,compsoc]{IEEEtran}
\usepackage{amsmath,amsfonts}
\usepackage{indentfirst}
\usepackage{graphicx,subfig}
\usepackage{makecell}
\usepackage{float}
\usepackage{amsmath}%
\usepackage{caption}
\usepackage{xcolor}
\usepackage{soul}
\usepackage{booktabs}

\usepackage[numbers,sort&compress]{natbib}
\usepackage[pagebackref=true,breaklinks=true,colorlinks,bookmarks=false]{hyperref}
\usepackage{multirow}
\begin{document}
\title{Light Field Depth Estimation via Stitched Epipolar Plane Images}
\author{Ping Zhou,
        Langqing Shi,
        Xiaoyang Liu,
        Jing Jin,
        Yuting Zhang,
       and 
        Junhui Hou, \emph{Senior Member, IEEE}
\IEEEcompsocitemizethanks{\IEEEcompsocthanksitem P. Zhou, L. Shi and X. Liu are with the School of Biological Science $\&$ Medical Engineering, Southeast University, NanJing, China. \protect\\
E-mails: capzhou@163.com; xyliu20@seu.edu.cn; yt.zhang1@outlook.com
\IEEEcompsocthanksitem J. Jin and J. Hou are are with the Department of Computer Science, City University of Hong Kong, Hong Kong, and also with the City University of Hong Kong Shenzhen Research Institute, Shenzhen 518057, China. \protect\\
E-mails: jingjin25-c@my.cityu.edu.hk; jh.hou@cityu.edu.hk
\IEEEcompsocthanksitem Junhui Hou and Ping Zhou are corresponding authors. This work was supported in part by the National Natural Science Foundation of China under Grants 52071075 and 11572087, in part by the Hong Kong Research Grants Council
under Grant 11218121, in part by Hong Kong Innovation and Technology
Fund under Grant MHP/117/21, and in part by the Basic Research General Program of Shenzhen Municipality under Grant JCYJ20190808183003968.

}
}
\maketitle

\begin{abstract}

Depth estimation is a fundamental problem in light field processing. Epipolar-plane image (EPI)-based methods often encounter challenges such as low accuracy in slope computation due to discretization errors and limited angular resolution. Besides, existing methods perform well in most regions but struggle to produce sharp edges in occluded regions and resolve ambiguities in texture-less regions. To address these issues, we propose the concept of stitched-EPI (SEPI) to enhance slope computation. SEPI achieves this by shifting and concatenating lines from different EPIs that correspond to the same 3D point. Moreover, we introduce the half-SEPI algorithm, which focuses exclusively on the non-occluded portion of lines to handle occlusion. Additionally, we present a depth propagation strategy aimed at improving depth estimation in texture-less regions. This strategy involves determining the depth of such regions by progressing from the edges towards the interior, prioritizing accurate regions over coarse regions. Through extensive experimental evaluations and ablation studies, we validate the effectiveness of our proposed method. The results demonstrate its superior ability to generate more accurate and robust depth maps across all regions compared to state-of-the-art methods.
The source code will be publicly available at \url{https://github.com/PingZhou-LF/Light-Field-Depth-Estimation-Based-on-Stitched-EPIs}.
\end{abstract}

\section{Introduction}\label{sec:introduction}
 
\IEEEPARstart{T}{he} light field (LF) is a high-dimensional function that describes the light rays permeating the 3D free space. In comparison to conventional 2D imaging, the LF image simultaneously captures spatial and angular information of light rays. This capability enables a variety of applications, including post-refocusing \cite{lfapp2014refocus}, 3D reconstruction \cite{lfdepth2013scene}, saliency detection \cite{lfapp2014saliency}, and virtual/augmented reality \cite{lfapp2015vrdisplay,lfapp2017vryu}, and so on.


Depth estimation is a critical challenge in LF image processing, as the accuracy of depth estimation greatly affects the performance of subsequent LF-based applications. Numerous depth estimation algorithms have been developed, leveraging the unique properties of LF images. These include stereo matching across sub-aperture images \cite{RN106,RN107,RN105,RN104,RN108,RN109}, plane-sweeping \cite{Chuchvara}, and the utilization of defocus and correspondence cues \cite{RN111,RN112,RN113,RN110}. Another popular approach is constructing epipolar-plane images (EPIs) \cite{bolles1987epipolar} from LF images for depth estimation. By observing that the projections of a given scene point in different sub-aperture images form straight lines in EPIs, the depth map can be derived by computing the slopes of these EPI lines \cite{RN120,RN119,RN123,RN121,khan2021edgeaware}. Despite the significant advancements achieved by existing EPI-based methods in LF depth estimation accuracy, several challenges remain unresolved. One challenge is the error introduced during slope computation. Since the slope computation from discrete points in EPIs involves a many-to-one mapping problem, multiple straight lines with similar slopes can generate the same set of discrete points. Consequently, uncertainty and ambiguity arise during slope computation, leading to reduced accuracy. This issue can be mitigated by employing diffusion methods \cite{khan2021edgeaware} for refinement. Another challenge stems from the inherent trade-off between angular and spatial resolutions, particularly the limited angular resolution in LF bandwidth products. This limitation imposes restrictions on the accuracy of depth estimation due to the sparse sampling of lines in EPIs. To address this limitation, angular super-resolution techniques for LF data \cite{RN133,RN132,RN135,RN134} can be employed as a pre-processing step. However, this introduces additional computational overhead.

The estimation of depth in texture-less regions remains a challenging issue due to the lack of distinguishing features. Most state-of-the-art algorithms address this problem using a global or local depth optimization framework, employing various penalization, smoothing, and reprojection strategies over texture-less regions. However, these approaches often fail to produce accurate depth measurements and result in ambiguities and artifacts \cite{RN59,RN53,RN145,RN51,Lourenco,khan}. 
Furthermore, the depth estimation over occluded regions is prone to errors, leading to the loss of fine structures due to color inconsistency in partially occluded regions. Although some methods have been developed to tackle occlusion, such as modified angular photo-consistency for simple occlusion \cite{wang}, a complete model for multi-occlusion \cite{RN138}, and partial focal stacks \cite{RN140}, the occlusion problem still poses significant challenges. Therefore, it is crucial to develop algorithms that can further improve depth estimation in both occluded and texture-less regions.
To tackle the aforementioned challenges, we present a new LF depth estimation framework. First, we quantitatively analyze the uncertainty of line slope computation using a discretization model for straight lines. Based on this analysis, we propose the concept of stitched-EPI (SEPI) by shifting and concatenating all EPIs corresponding to the same scene point. The SEPI encompasses a greater number of projections compared to the commonly-used EPI, thereby enhancing the accuracy of LF depth estimation. Additionally, we introduce effective algorithms to address challenges related to occlusion and texture-less regions, respectively. Extensive experiments over both benchmark datasets and real world LF images demonstrate the advantage of our method over state-of-the-art methods. 

We summarize the main contributions of this paper as follows:


\begin{itemize}

\item We propose the SEPI representation for LF depth estimation, based on the theoretical discretization model of straight lines. Besides, we propose half-SEPI to handle the occlusion issue.

\item We propose a depth propagation strategy to deal with LF depth estimation over texture-less regions.
\end{itemize}


The remainder of the paper is structured as follows: In Section \ref{sec:RW}, we provide a concise review of previous studies on LF depth estimation. Following this, in Section \ref{sec:S-EPI}, we present the SEPI representation, and in Section \ref{sec:SEPI depth}, we provide a comprehensive explanation of our SEPI-based depth estimation algorithm. We showcase the experimental results and perform comparisons with other state-of-the-art methods in Section \ref{sec:exp}. Finally, we conclude this paper in Section \ref{sec:con}.

\section{Related Work}
\label{sec:RW}

The existing LF depth estimation methods can be roughly classified into four categories: 
the matching-based methods, the cues-based methods, the EPI-based methods, and the learning-based methods.

\subsection{Matching-based Methods}
Different constraints are contained in the matching-based methods instead of using traditional stereo-matching methods. 
Heber \emph{et al}. \cite{RN150} estimated depth by matching the central view with other sub-aperture images, although not utilizing all sub-aperture image pairs. To enhance depth estimation, Heber \emph{et al}. \cite{RN63} further introduced a novel principal component analysis (PCA) technique to align sub-aperture images, transforming the depth estimation problem into a rank-minimization problem. Jeon \emph{et al}. \cite{jeon} accurately estimated sub-pixel shifts of sub-aperture images by applying the phase shift theorem in the Fourier domain. Yucer \emph{et al}. \cite{RN56} proposed the LF gradient method to locally match patches between adjacent sub-aperture images. Due to the narrow baseline, stereo-matching methods inevitably involve interpolation, leading to uncertain and ambiguous depth estimation.

\subsection{Cues-based Methods}

Significant efforts have been dedicated to depth estimation, utilizing various cues. Ng \emph{et al}. \cite{RN142} demonstrated the possibility of refocusing an LF by rearranging the light rays. Tao \emph{et al}. \cite{tao} proposed a fusion method that combines defocus and correspondence cues to estimate depth, further enhancing it using the normal map and exploring the shading cue. Williem \emph{et al}. \cite{RN55} introduced an adaptive defocus cue and angular entropy to evaluate the angular image's randomness for depth estimation. Lin \emph{et al}. \cite{RN58} constructed a focal stack and employed color symmetry to locate the optimal depth. For handling occlusion, Chen \emph{et al}. \cite{RN62} proposed a bilateral consistency method to determine the likelihood of occlusion and subsequently refined the depth of points with high occlusion possibility. Wang \emph{et al}. \cite{wang} developed a modified angular photo-consistency approach specifically designed for simple occlusions. Additionally, Zhu \emph{et al}. \cite{RN51} formulated a comprehensive model to address complex and intricate occlusion scenarios.

\subsection{EPI-based Methods}

The EPI plays a crucial role in depth estimation. Bolles \emph{et al}. \cite{RN67} introduced the concept of EPI and applied it to 3D reconstruction based on camera motion theory. Wanner \emph{et al}. \cite{RN61} proposed the use of structure tensor for local line orientation estimation. To enhance the performance of the structure tensor in scenarios with large disparities, Suzuki \emph{et al}. \cite{RN54} employed EPI shearing to align the lines vertically before applying the structure tensor. Additionally, Ziegler \emph{et al}. \cite{RN66} extended 2D EPI to the 4D EPI volume and 3D EPI space, respectively. To address occlusion, Zhang \emph{et al}. \cite{RN53} introduced a spinning parallelogram operator that divides the EPI into two parts, where the earth mover's distance is maximized to obtain the optimal orientation of the parallelogram for each point. 
Chen \emph{et al}. \cite{chen} handled the occlusion issue by manipulating the shrinkage or reinforcement weights over the partially occluded border region, which was detected with the superpixel-based regularization. 

\subsection{Learning-based Methods}

With the advancements in deep learning, numerous convolutional neural network (CNN)-based methods have been proposed \cite{RN130,RN127,RN124,RN125,RN129,RN52,jinjing,Shin,tsai}. Heber \emph{et al}. \cite{RN52} introduced a U-shaped auto-encoder-style network that utilizes the 3D EPI volume as input to estimate depth. Peng \emph{et al}. \cite{RN127} designed a loss function that combines compliance and divergence constraints to address missing information caused by warping. Shi \emph{et al}. \cite{Shin} proposed a fully-CNN based on light field geometry to overcome the problem of insufficient data. Wu \emph{et al}. \cite{RN124} developed a CNN that fuses sheared EPIs and learns the optimal shear value for depth estimation. Tsai \emph{et al}. \cite{tsai} introduced a view selection module that generates attention maps, effectively utilizing all views. Jin \emph{et al}. \cite{jinjing} proposed an unsupervised learning-based method that leverages the geometric structure of LF data and incorporates sub-LFs to enhance accuracy in occluded regions.

\begin{figure*}[h]
\centering
\includegraphics[width=18.1cm]{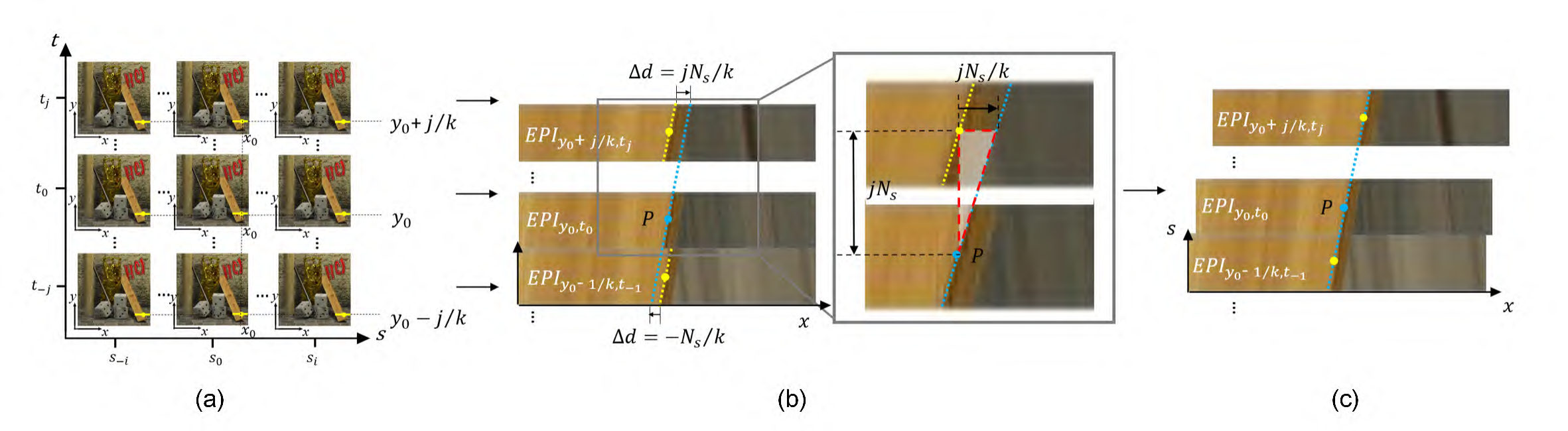}
\caption{Pipeline of the SEPI construction. (a) Sub-aperture images. (b) The corresponding lines of point \emph{P} in different EPIs. (c) EPIs are shifted and stitched to construct an SEPI.}
\label{fig:stitched_epi}
\end{figure*}

\section{Stitched-EPI Representation} 
\label{sec:S-EPI}


 The 4D LF image, denoted by \(L\left( x,y,s,t \right)\), is represented using the two-plane parameterization, where \(\left( x,y \right)\)  and \(\left( s,t \right)\) are the spatial and angular coordinates, respectively. EPIs, denoted as $E_{y,t}(x,s)$ or $E_{x,s}(y,t)$, are 2D slices of the LF image constructed by fixing two of the four dimensions. Under the Lambertian assumption, a typical scene point in 3D space is mapped to a straight line in an EPI, known as photo-consistency, with the slope of the line reflecting the distance between the scene point and the camera \cite{RN142}.
 Consequently, LF depth estimation involves estimating the line slopes of EPIs. However, since an LF image is acquired through discrete and finite sampling of the continuous ray-space, the lines formed in EPIs are discretizations of continuous straight lines. This discrete property poses challenges for accurate slope estimation, thus limiting the accuracy of EPI-based depth estimation methods. 
 
Based on our theoretical analysis in the \textit{Supplementary Material}, which is established upon a discretization model for straight lines, it is evident that a line in the EPI with a higher pixel count leads to more accurate depth estimation. Therefore, we propose the concept of stitched-EPI (SEPI) to achieve this objective by concatenating lines from different common EPIs that correspond to the same point in the scene.  
For a point $I_{s_0,t_0} (x_0, y_0)$ in the central view of an LF image, we assume its corresponding line slope is \emph{k}. 
In the absence of occlusions, there exist corresponding pixels of $I_{s_0,t_0} (x_0, y_0)$ in other sub-aperture images, and thus, there exist more EPIs containing straight lines with the same slope of \emph{k}. Specifically, for a row of sub-aperture images at the same angular position $t_j$, corresponding pixels of $I_{s_0,t_0} (x_0, y_0)$ share the same spatial coordinate at $y$ axis, i.e., $y_0 + j/k$, where $j = t_j-t_0$, as shown in Figure \ref{fig:stitched_epi} \textcolor{red}{(a)}. Therefore, we can construct some special 2D slices, i.e., the horizontal EPIs $E_{y_j,t_j}$, $y_j = y_0 + j/k$, and they contain discrete straight lines passing through $(x_0, s_0)$, as shown in Figure \ref{fig:stitched_epi} \textcolor{red}{(b)} by the yellow dotted lines. As $E_{y_j, t_j}$ and $E_{y_0, t_0}$ all contain the straight line related to the same point in 3D scene, we denote them as \textit{corresponding EPIs}. Note that the straight lines related to the same point in \textit{corresponding EPIs} are parallel to each other (see Figure \ref{fig:stitched_epi} \textcolor{red}{(b)}), and thus we first shift these lines and then concatenate them together to construct the SEPI: 
\begin{equation}
\text{SEPI}(x, s^{'}) = \mathop{\mathcal{C}}\limits_{j} \left( \mathcal{T} \left( E_{y_j,t_j} (x, s) \right) \right),
\label{equ:8}
\end{equation}
where $\mathcal{C}(\cdot)$ and $\mathcal{T}(\cdot)$ are the operators shifting and then concatenating $E_{y_j, t_j}$, 
as shown in Figure \ref{fig:stitched_epi} (c).
$s^{'}$ means that the angular resolution in \emph{s}-axis has changed after construction of SEPI. 
The shifting operator $\mathcal{T}(\cdot)$ is defined as:
\begin{equation}
\mathcal{T}\left( E_{y_j,t_j} (x, s) \right) = E_{y_j ,t_j}\left( x - \frac{jN_s}{k}, s \right),
\label{equ:9}
\end{equation}
where \(N_{s}\) represents the angular resolution in the \emph{s}-axis. As depicted in Eq. \eqref{equ:9}, the operator $\mathcal{T}(\cdot)$ shifts each pixel of $E_{y_j,t_j}$ by \(jN_{s}/k\) intervals along the \emph{x}-axis. In Figure \ref{fig:stitched_epi} \textcolor{red}{(b)}, it can be observed that the discrete lines in the shifted \textit{corresponding EPIs} are parallel to each other.
The shifting interval is the distance between these discrete lines in the \emph{x}-axis. Similarly, in the \emph{s}-axis, the distance between the shifted EPIs is the product of the angular resolution \(N_{s}\) and the index \emph{j} in the \emph{t}-axis of the EPI. Consequently, by leveraging the properties of a right triangle, the shifting interval can be calculated as \(jN_{s}/k\). 
A positive value of \(jN_{s}/k\)  indicates a rightward shift of the EPI, while a negative value implies a leftward shift. 
Subsequently, the operator $\mathcal{C}(\cdot)$ concatenates lines in the shifted EPIs, sorted by $j$, along the \emph{s}-axis. When the candidate \emph{k} matches the desired slope \(\widehat{k}\), the lines from different EPIs are combined into a new straight line in the SEPI, as demonstrated by the blue dotted line in Figure  \ref{fig:stitched_epi} \textcolor{red}{(c)}. 

In SEPI, the resolution of the SEPI in \emph{x}-axis is identical to that of the traditional EPI, but the resolution of the SEPI in \emph{s}-axis is \(N_{t}\) times their original resolution, which is essential to improve the initial depth estimation result in Section \ref{sec:dis0}.

\begin{figure*}[ht]
\centering
\includegraphics[width=18.1cm]{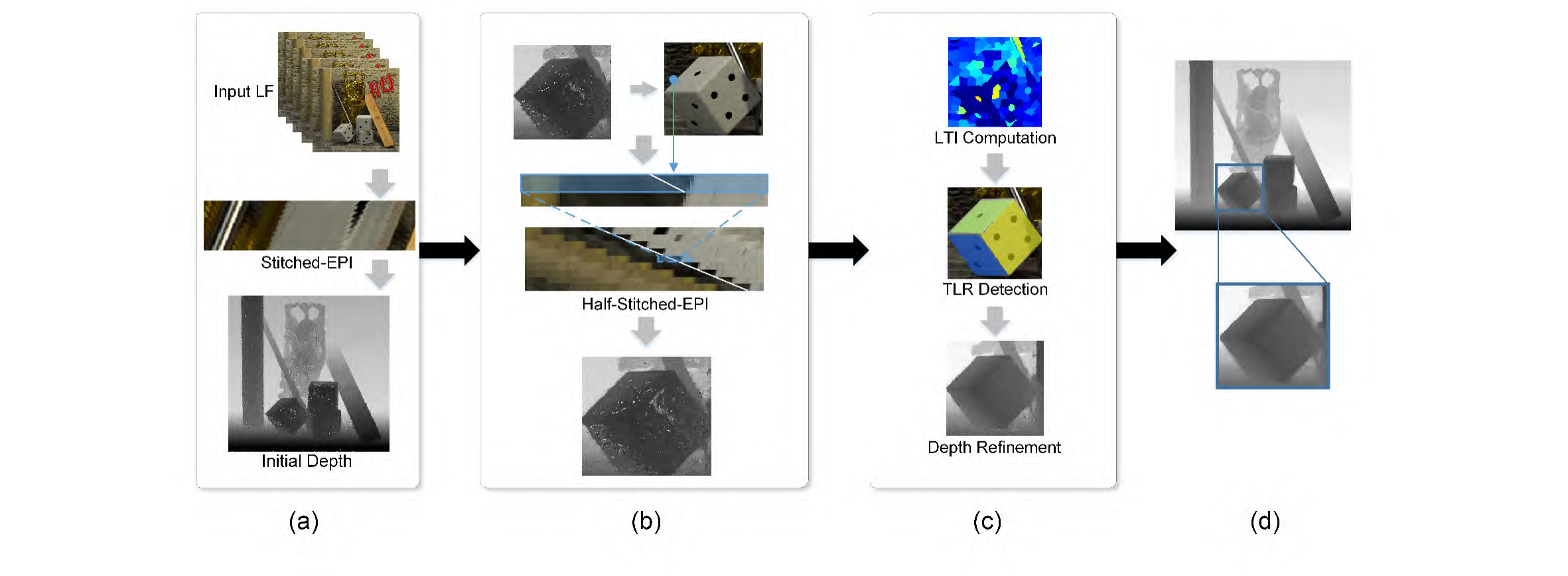}
\caption{Pipeline of the proposed LF depth estimation method. (a) SEPI-based initial depth estimation. (b) Half-SEPI-based depth refinement over occluded regions. (c) Depth refinement over texture-less regions. (d) Global optimization.}
\label{fig:pipline}
\end{figure*}

\section{Proposed LF Depth Estimation}
\label{sec:SEPI depth}

\textbf{Overview}. As illustrated in Figure \ref{fig:pipline}, our LF depth estimation method consists of the following four modules: 
\begin{enumerate}
\item[1)] \emph{SEPI-based initial depth  estimation.} 
We first compute the slope map using the SEPI algorithm. For each point in the central sub-aperture image, an SEPI is constructed, allowing us to obtain a more precise slope map. The increased number of pixels in the lines of SEPIs contributes to this enhanced accuracy.
(see Section \ref{sec:dis0})

\item[2)] \emph{Half-SEPI-based depth refinement over occluded regions.} To improve initial slope results over occluded regions, we propose half-SEPI (half-SEPI) algorithm that predicts, shifts, and concatenates non-occluded points in corresponding EPIs. 
(see Section \ref{sec:occRefine})

\item[3)] \emph{Depth refinement over texture-less regions.} 
Initially, we distinguish texture-less regions in the LF image using local texture information (LTI) and the color constraint at the superpixel scale. Subsequently, we refine the coarse depth by propagating accurate depth information from the edges to the interior of the texture-less regions.
(see Section \ref{sec:disTlm})

\item[4)] \emph{Global depth optimization.} A series of rectification and reinforcement operations are performed and fitted into the global optimization model. 
(see Section \ref{sec:disFinal})
    
\end{enumerate}
In what follows, we will detail these modules one by one.

\begin{figure}[!t]
\centering
\includegraphics[width=8.8cm]{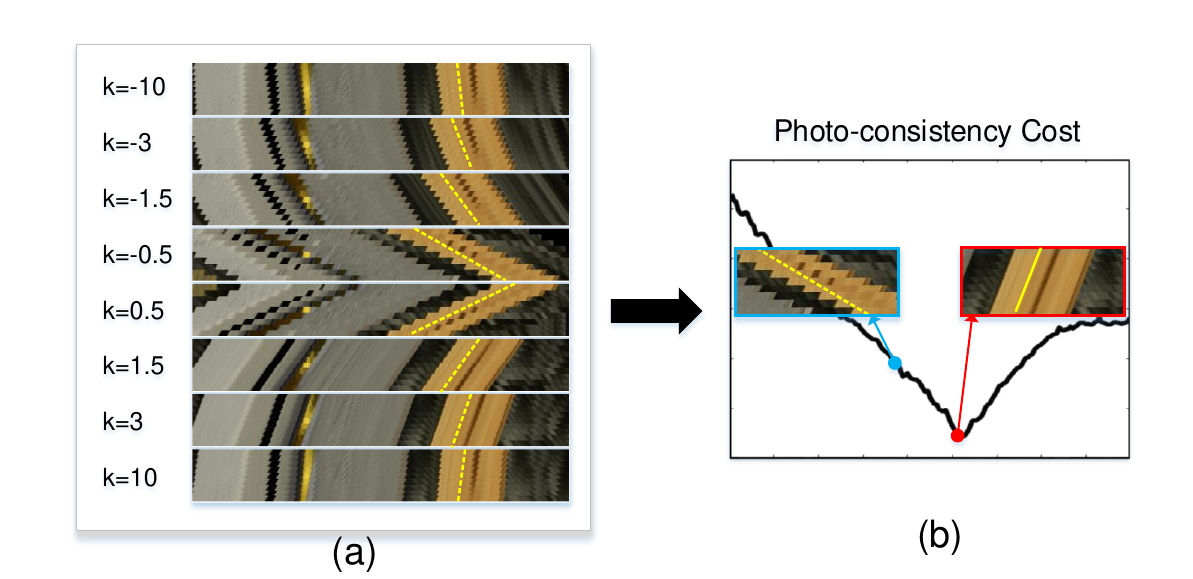}
\caption{SEPI-based initial slope computation. (a) SEPIs with different candidate \emph{k} values. (b) the color variance along the straight line determined by \emph{k}, and the initial depth value is determined by the minimum color variance.}
\label{fig:slopeRefine}
\end{figure}

\subsection{SEPI-based Initial Depth Estimation}
\label{sec:dis0}


As discussed in Section \ref{sec:S-EPI}, the construction of SEPI involves shifting and stitching corresponding EPIs. It is important to note that the shifting interval depends not only on the angular coordinate \emph{j} but also on the candidate slope \emph{k}. Figure \ref{fig:slopeRefine} \textcolor{red}{(a)} illustrates that when the candidate slope \emph{k} matches the desired slope during the shifting operation, the concatenated line's pixels maintain photo-consistency (indicated by the yellow line in the red square). Conversely, when the candidate slope differs from the desired slope, the photo-consistency is disrupted along the discrete line (represented by the yellow dotted line).


Hence, for each pixel in the central sub-aperture image, multiple SEPIs are constructed with different candidate slopes \emph{k}. The color variance $V_{x_{0},y_{0}}$ along the direction determined by \emph{k} in the corresponding SEPI is computed to assess photo-consistency, as shown in Equation (\ref{equ:10}): 
\begin{equation}
V_{x_{0},y_{0}}\left( k \right) = \sum_{\left( x,s \right)\mathbf{\in}l}^{}\left( \text{SEPI}_{k}\left( x,s \right) - I_{s_0,t_0} (x_0, y_0) \right)^{2},
\label{equ:10}
\end{equation} 
where  \(\text{SEPI}_{k}(x,s)\) refers to the SEPI with slope \emph{k}, and \(l\) is the line with candidate slope \(k\). Therefore, as shown in Figure \ref{fig:slopeRefine} \textcolor{red}{(b)}, a series of color variance values are obtained according to candidate \emph{k} values, and the initial slope for $I_{s_0,t_0} (x_0, y_0)$ is determined by:

\begin{equation}
\widehat{k} = \mathop{\arg\min}_{k}\operatorname{}{V_{x_{0},y_{0}}\left( k \right)}.
\label{equ:11}
\end{equation}
Considering the disparity range of LF datasets \cite{wanner} and \cite{RN148}, we consider 181 distinct values for the candidate slope \emph{k}.
Besides, we define the confidence as the ratio of the mean variance to minimum variance among all candidate slopes:
\begin{equation}
c_{x_{0},y_{0}} = \frac{\text{mean}\left( V_{x_{0},y_{0}}\left( k \right) \right)}{\min\left( V_{x_{0},y_{0}}\left( k \right) \right)}, 
\label{equ:12}
\end{equation}
which is assigned to the initial slope.

\textbf{Remark}. 
Within a single EPI, a discrete line encompasses only \(N_{s}\) (or \(N_{t}\)) pixels, representing the angular resolution along the \(s\)-axis (or\(\ t\)-axis). In other words, only rays from a subset of sub-aperture images are taken into account, leading to inaccuracies and ambiguities in the results. However, in the case of the SEPI, a discrete line comprises \(N_{s}N_{t}\) pixels, allowing for the inclusion of rays from a larger number of sub-aperture images in the slope computation. Consequently, the range of uncertain slopes is significantly narrowed down, resulting in improved accuracy in the slope computation.

\subsection{Half-SEPI-based Depth Refinement over Occluded Regions}
\label{sec:occRefine}

\begin{figure}[!t]
\centering
\includegraphics[width=8.8cm]{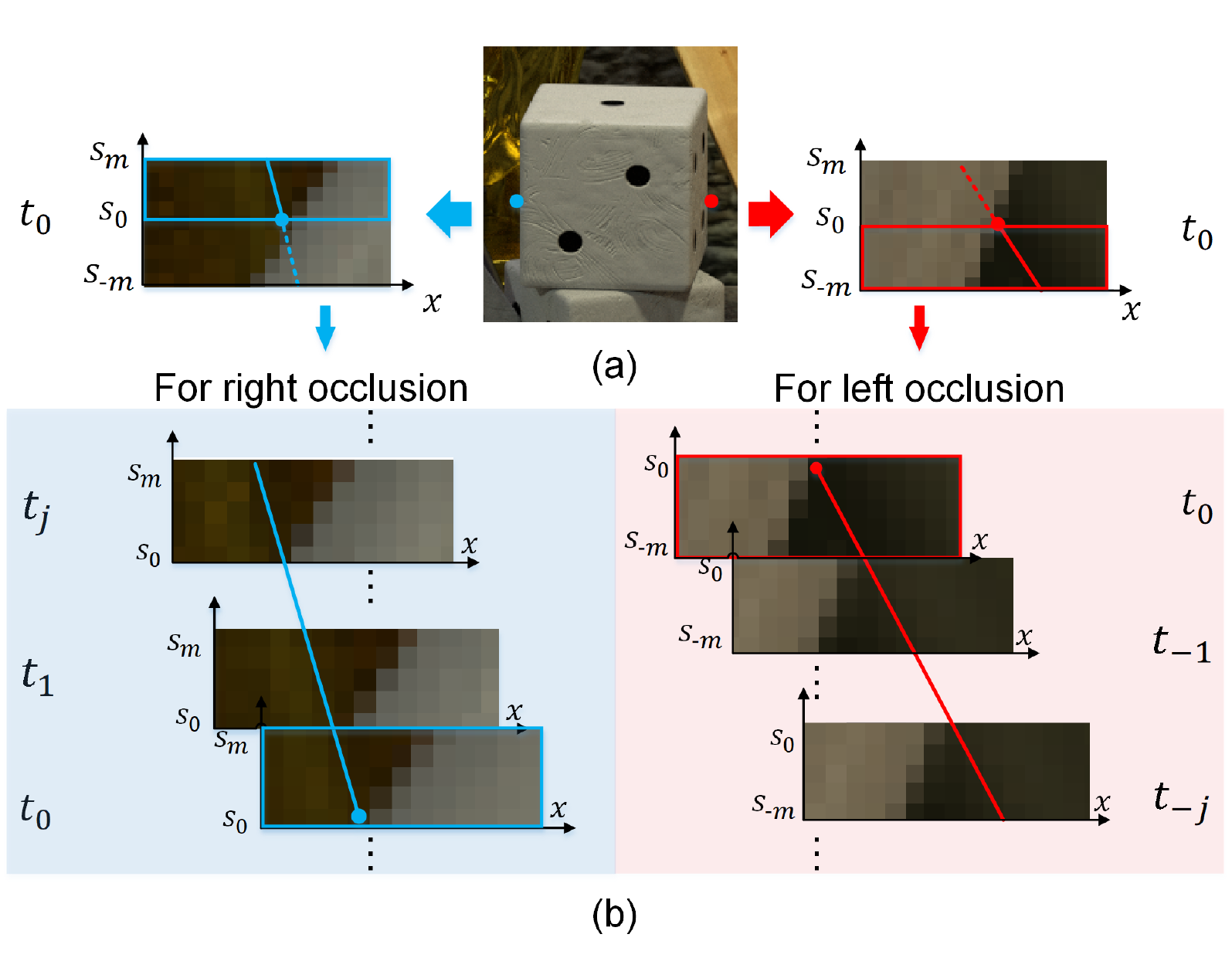}
\caption{Illustration of Half-SEPI (horizontal occlusion). (a) The right and left occlusion. (b) Half-SEPIs corresponding to right and left occlusion.}
\label{fig:occlusion}
\end{figure}

Considering the inherent occlusion problem that is not accounted for during the construction of SEPIs, the depth estimation module based on SEPIs, as described in the previous section, may yield errors in occluded regions. To tackle this challenge, we present the half-SEPI algorithm, which exclusively shifts and concatenates non-occluded pixels from different EPIs, as illustrated in Figure \ref{fig:occlusion}.

Using the initial depth maps obtained through the SEPI approach, we employ mean-shift clustering on the patch of the central sub-aperture image and the initial depth map to refine the occluded regions calculated by the angular model in \cite{wang}. Firstly, we apply Canny edge detection to the central view image. Subsequently, an edge orientation predictor is utilized to determine the orientation angles at each edge pixel, which serves as candidate occlusion pixels in the central view. Following this, the half-SEPI approach is applied to refine inaccurate slope results in the occluded regions. The determination of how to shift and concatenate non-occluded points in different EPIs is critical for the half-SEPI algorithm and depends on the directional relationship between the foreground and background. In this study, we classify occlusion into two categories: horizontal (left, right) and longitudinal (up, down) occlusion. For instance, in the case of horizontal occlusion, the blue and red points in Figure \ref{fig:occlusion} \textcolor{red}{(a)} represent right and left occlusions, respectively. Additionally, the solid and dotted lines in Figure \ref{fig:occlusion} \textcolor{red}{(a)} are constructed using non-occluded and occluded pixels with their respective desired slopes.

For a point \emph{p}\((x_0,y_0)\) ($I_{s_0,t_0} (x_0, y_0)$) in central sub-aperture image, \emph{p}\(_L(x_0-4,y_0)\) represents the point to its left, and \emph{p}\(_R(x_0+4,y_0)\) represents the point to its right. We obtain the average depths of their $3\times 3$ adjacent areas and denote them as \(D_p\), \(D_L\), and \( D_R\) for \(p\), \(p_L\), and \(p_R\), respectively. If \(D_p<D_R \) and \(D_L<D_R\) hold, then \(p\) is classified as a left occlusion. Conversely, if \(D_p<D_L\) and \(D_R<D_L\) hold, \(p\) is classified as a right occlusion. Therefore, we select sub-aperture images for constructing the half-SEPI based on the occlusion categorization. If \(p\)  represents a left occlusion, we choose only the sub-aperture images to the right of the central sub-aperture image; otherwise, we choose only the sub-aperture images to the left


With half-SEPIs, the similar slope computation in Section \ref{sec:dis0} is performed for occluded regions to refine the initial depth map obtained by the SEPI approach.

\subsection{Depth Refinement over Texture-less Regions}
\label{sec:disTlm}

\begin{figure}[!t]
\centering
\includegraphics[width=8.8cm]{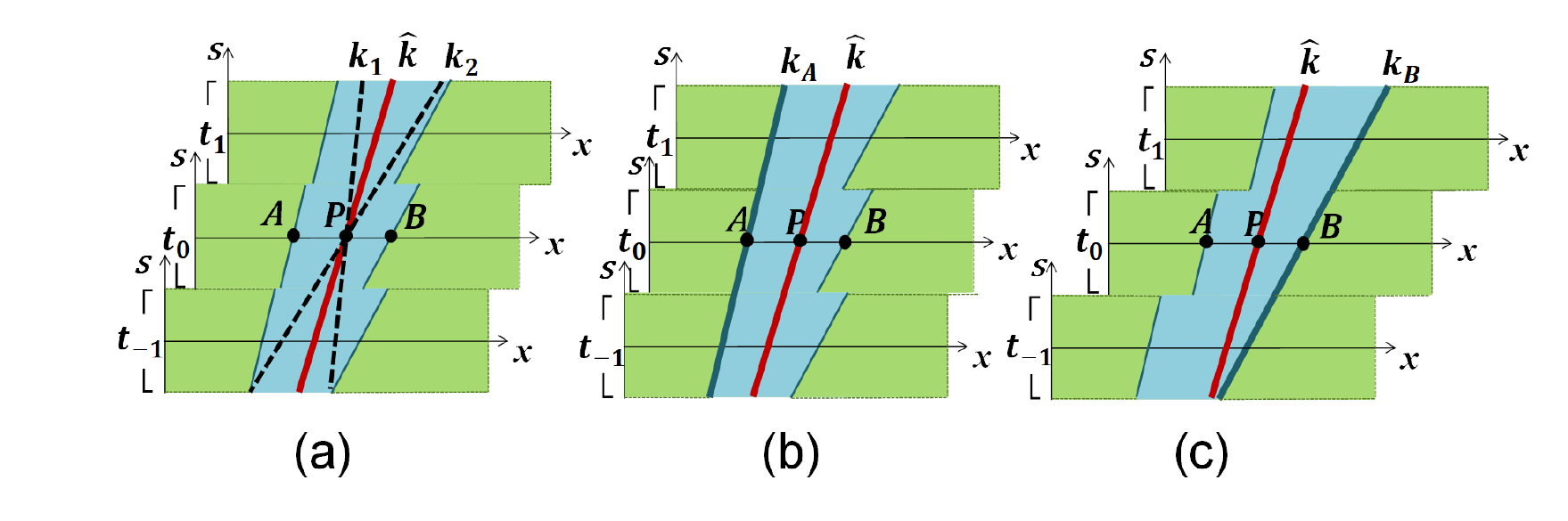}
\caption{Illustration of ambiguity over a texture-less region. (a) \(P\) is in the texture-less region. (b) Point \emph{A} on the left edge of the texture-less region. (c) Point \emph{B} on the right edge of the texture-less region.}
\label{fig:TLR}
\end{figure}

The preceding three modules primarily focus on improving the accuracy of slope estimation for regions with abundant texture information. However, when it comes to texture-less regions, the slopes obtained using these modules remain unreliable. To illustrate this point, let's consider a point \(p\) situated in a texture-less region, with \(\widehat{k}\) representing its true slope. As depicted in Figure \ref{fig:TLR} \textcolor{red}{(a)}, the photo-consistency property holds for all lines with slopes ranging between \(k_{1}\) and \(k_{2}\). Consequently, if the SEPI is constructed using the ideal slope \(\widehat{k}\), the lack of texture unavoidably results in ambiguity. 
As depicted in Figures \ref{fig:TLR} \textcolor{red}{(b)} and \textcolor{red}{(c)}, accurate slope computation within texture-less regions poses a challenge \cite{RN59,RN51,RN53,RN145}. However, the edge lines \(l_{A}\) and \(l_{B}\) in these regions exhibit high local contrast, offering the potential for relatively precise slope computation. Leveraging this observation, we refine the line slope within texture-less regions by propagating the slope from the edges towards the interior, utilizing information from accurate regions to improve results in coarse regions.

\begin{figure}[!t]
\centering
\includegraphics[width=8.8cm]{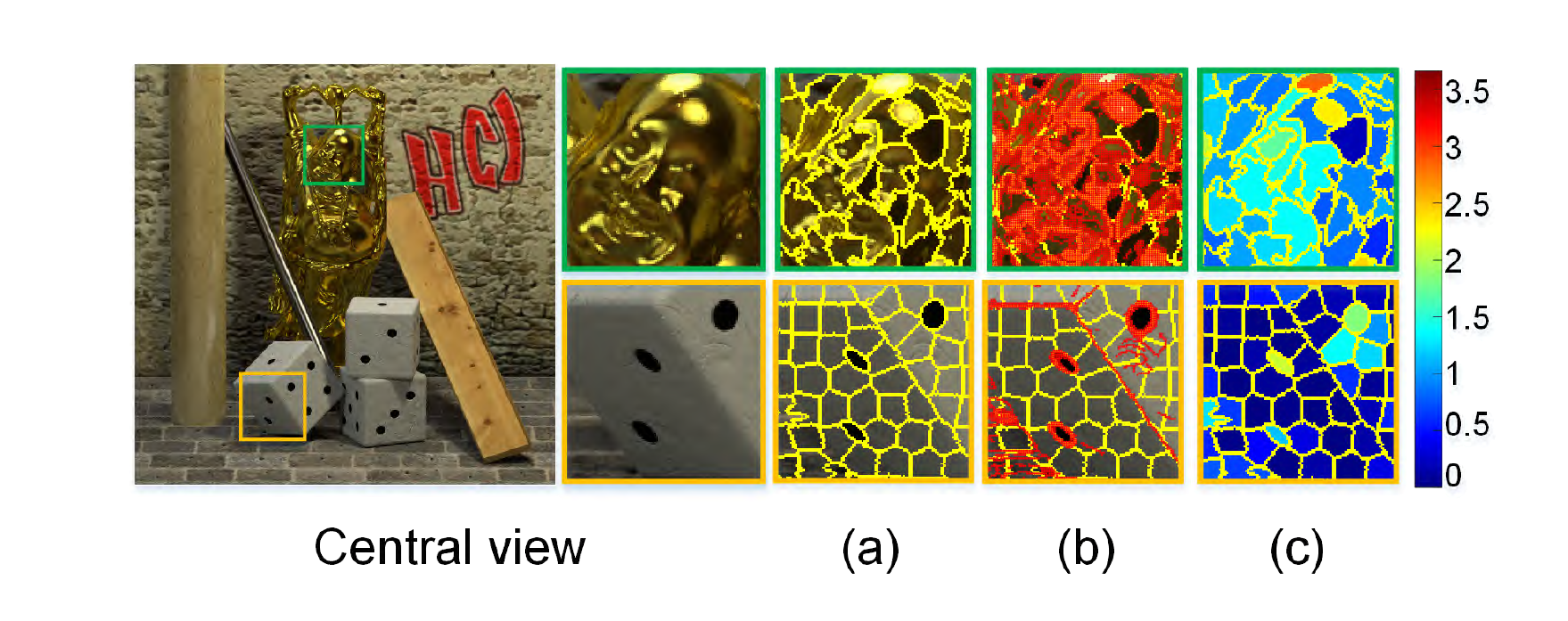}
\caption{Texture pixels and LTI for the LF data \emph{buddha2}.  (a) Superpixel boundaries (overlaid in yellow). (b) Texture pixels (marked with red dots). (c) Close-up LTI maps}
\label{fig:LTI}
\end{figure}

Specifically, we use the superpixel derived by \cite{RN144} as the smallest texture-less region unit, considering that superpixel segmentation usually groups similar pixels while maintaining consistency with the underlying texture. 
To assess whether a superpixel qualifies as a texture-less region, we introduce the concept of Local Texture Information (LTI). Initially, texture points are identified based on the results obtained from slope computation (as described in Section \ref{sec:occRefine}) and Canny edge detection. For a given superpixel \(\text{SP}_{i}\), its LTI is defined as 
\begin{equation}
\text{LTI}_{i} = \sum_{j = 1}^{n_{i}}\frac{\delta_{\text{ij}}\text{SP}_{i,j}}{n_{i}},
\label{equ:18}
\end{equation}
where\(\text{\ SP}_{i,j}\) is the \(jth\) pixel in superpixel \(\text{SP}_{i}\),
\(n_{i}\) is the number of pixels in \(\text{SP}_{i}\) , and
\(\delta_{ij}\)=1 only if pixel \(\text{\ SP}_{i,j}\) is
considered as a texture point and not edge of \(\text{SP}_{i}\). 

The LTI serves as an indicator of a superpixel's likelihood of being a texture-less region, with lower LTI values indicating a higher probability. To determine the texture-less regions, we employ an adaptive thresholding approach. Figure \ref{fig:LTI} illustrates two close-up views of the light field image \emph{Buddha2}, where the majority of superpixels correspond to textured regions in the face region, whereas nearly all superpixels in the dice image represent texture-less regions.


The texture-less region identified through the superpixel method is typically small \cite{RN144}, resulting in a linear or piecewise linear depth variation within a larger nonlinear region. It should be noted that the piecewise linear model is commonly assumed in various methods. Therefore, we employ a slope linear propagation strategy within each texture-less region. Additionally, we consider the continuity of depth between adjacent texture-less regions. If adjacent superpixels belong to the same linear texture-less region, their normal vectors exhibit similarity. Conversely, neighboring superpixels with different normal vectors may belong to a larger nonlinear texture-less region. To determine whether there is linear depth change between texture-less regions, we utilize the HSV color space values to establish a propagation cue based on the mean-shift algorithm \cite{RN146}. Consequently, for all superpixels identified as texture-less regions, the depth is refined by minimizing the following energy function:
\begin{equation}
 \sum_{i}^{}\left( \left| \widehat{d}\left( \text{SP}_{i} \right) - d\left( \text{SP}_{i} \right) \right| +\tau_{\text{out}}\sum_{j}^{}\frac{\left| \widehat{d}\left( \text{SP}_{j} \right) - d_{l}\left( \text{SP}_{j} \right) \right|}{\nabla v \cdot w_{s}} \right),   
 \label{equ:19}
\end{equation}
where\(\widehat{\text{\ d}}\) is the refined depth, \(d\) is the depth result obtained in Section \ref{sec:occRefine}, \(d_{l}\) is the depth derived after slope linear propagating above, \(\text{SP}_{j}\) is the adjacent superpixel of \(\text{SP}_{i}\), \(\tau_{\text{out}}\) is the weight of the depth continuity between texture-less regions, \(w_{s}\) is the spatial distance between adjacent superpixels, and \(\nabla v\) is the gradient of mean depth in adjacent SPs. 


The first term in Eq. (\ref{equ:19}) enforces the refined depth to closely resemble the initial depth, while the second term ensures the refined depth aligns with its neighboring values. The denominators in the equation reduce the strength of the effect when the product of the gradient and spatial distance between adjacent superpixels is large.

\begin{figure*}[!t]
\centering
\includegraphics[width=18.1cm]{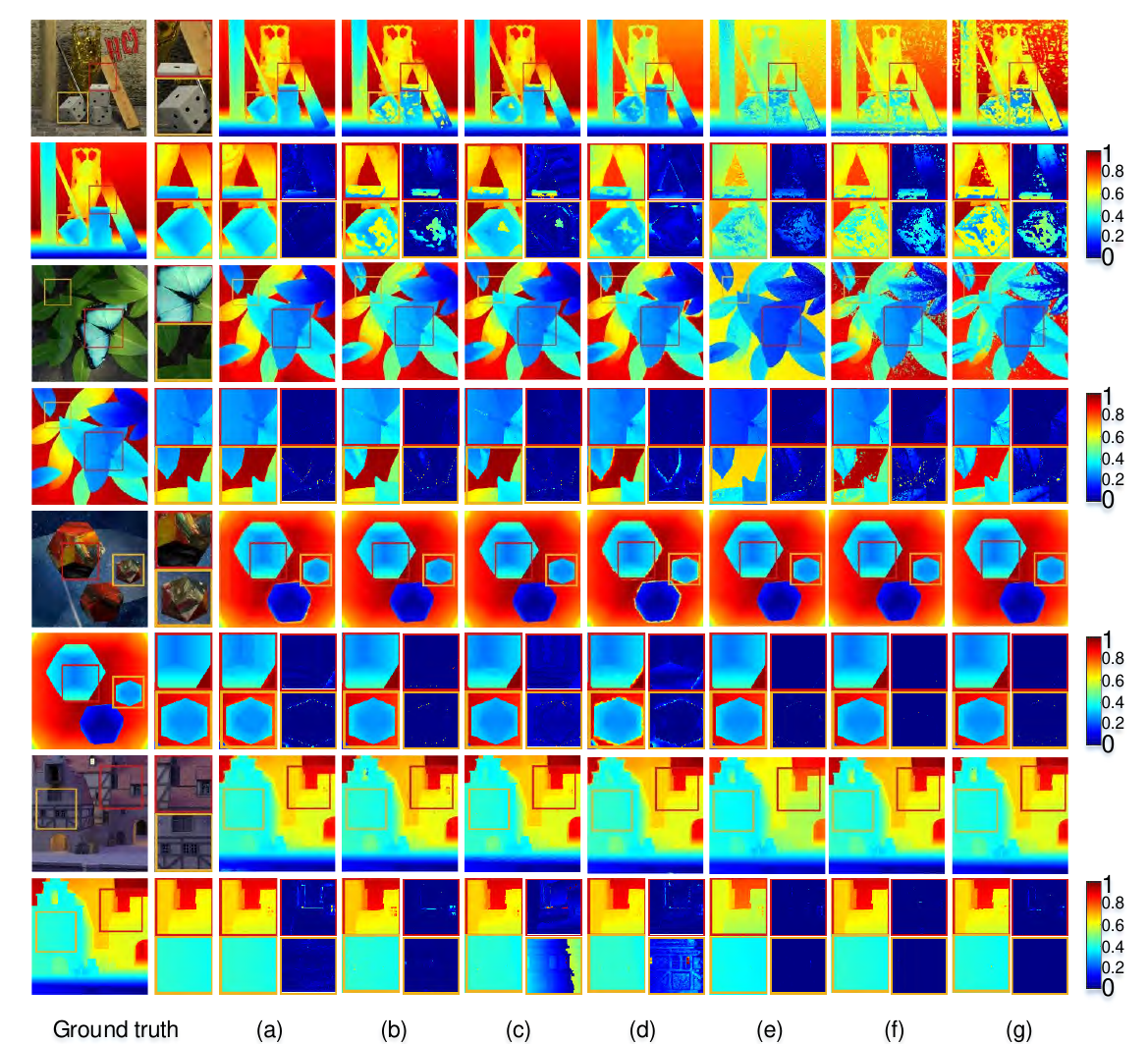}
\caption{Visual comparison of estimated depth maps on synthetic LF data  \emph{buddha2}, \emph{papillon}, \emph{platonic} and \emph{medieval2}. (a) Ours; (b) Han \emph{et al}. \cite{han};(c) Williem \emph{et al}. \cite{williem}; (d) Chen \emph{et al}. \cite{chen}; (e) Wang \emph{et al}. \cite{wangL}; (f) Tsai \emph{et al}. \cite{tsai}; (g) Shin \emph{et al}. \cite{Shin}. Note that (e), (f) and (g) are learning-based methods. The central sub-aperture image,
the close-up views of two representative regions and their ground-truth
depth maps are illustrated on the left column. The top row shows the
depth maps by different methods, and the bottom row shows the
depth maps and the error maps of the close-up views .}
\label{fig:monas}
\end{figure*}

\begin{table*}[!t]\small
\centering
\caption{Quantitative comparisons (100$\times$ MSE) of different methods on HCI Blender \cite{wanner} and HCI LF benchmark dataset \cite{RN148}.  (a) Ours, (b) Han \emph{et al}. \cite{han}, (c) Williem \emph{et al}. \cite{williem}, (d) Chen \emph{et al}. \cite{chen}, (e) Zhang \emph{et al}. \cite{zhang}, (f) Mishiba \emph{et al}. \cite{Mishiba}, (g) Jeon \emph{et al}. \cite{jeon}, (h) Wang \emph{et al}. \cite{wangL}, (i) Tsai \emph{et al}. \cite{tsai}, (j) Shin \emph{et al}. \cite{Shin}, (k) Jin \emph{et al}. \cite{jinjing}. The best and second best results of non-learning-based methods are highlighted in \textcolor{red}{red} and \textcolor{blue}{blue}, respectively. 16 scenes (from Antinous to Vinyl) are commonly adopted as the training set for learning-based methods, so we mark them as '-'.}
\begin{tabular}{@{}l|lllllll|llll@{}}
\toprule
& \multicolumn{7}{c|}{Non-learning-based}        & \multicolumn{4}{c}{Learning-based} \\
 &
  (a)&
  (b)&
  (c)&
  (d)&
  (e)&
  (f)&
  (g)&
  (h)&
  (i)&
  (j)&
  (k)\\ \midrule
Boxes &
  \textcolor{red}{5.20}  &
  \textcolor{blue}{6.70} &
  10.93 &
  9.62 &
  8.29 &
  11.43 &
  25.55 &
  2.66 &
  3.55 &
  5.90 &
  6.59 \\
Cotton &
  \textcolor{blue}{1.20} &
  1.24 &
  4.17 &
  5.90 &
  1.94 &
  \textcolor{red} {0.91} &
  13.59 &
  0.21 &
  0.22 &
  0.28 &
  1.99 \\
Dino &
   0.40 &
  \textcolor{red}{0.25}  &
  0.55 &
  0.95 &
  \textcolor{blue}{0.29} &
  0.62 &
  1.34 &
  0.25 &
  0.08 &
  0.17 &
  0.93 \\
Sideboard &
   1.02 &
  \textcolor{blue}{0.99} &
  2.16 &
  1.59 &
  \textcolor{red} {0.95} &
  1.80 &
  10.95 &
  0.76 &
  0.48 &
  0.85 &
  1.88 \\
Antinous &
  12.84 &
  20.19 &
  69.95 &
  53.28 &
  \textcolor{blue} {10.26} &
  \textcolor{red} {5.25} &
  112.61 &
  - &
  - &
  - &
  - \\
Boardgames &
  \textcolor{blue}{0.26} &
  0.41 &
  0.87 &
  \textcolor{red} {0.20} &
  0.53 &
  1.26 &
  6.78 &
  - &
  - &
  - &
  - \\
Dishes&
  1.29 &
  \textcolor{red} {0.93} &
  65.80 &
  2.60 &
  \textcolor{blue}{1.07} &
  1.52 &
  7.83 &
  - &
  - &
  - &
  - \\
Greek &
  \textcolor{red} {14.20} &
  51.09 &
  120.58 &
  107.03 &
  \textcolor{blue}{39.27} &
  83.44 &
  164.57 &
  - &
  - &
  - &
  - \\
Kitchen &
  6.77 &
  11.48 &
  12.67 &
  \textcolor{blue}{6.43} &
  10.83 &
  6.83 &
  \textcolor{red} {6.04} &
  - &
  - &
  - &
  - \\
Medieval2 &
  \textcolor{blue}{0.49} &
  \textcolor{red} {0.39} &
  1.07 &
  0.64 &
  0.56 &
  1.08 &
  4.69 &
  - &
  - &
  - &
  - \\
Museum &
  4.09 &
  3.84 &
  7.59 &
  \textcolor{blue}{3.64} &
  \textcolor{red} {2.31} &
  9.19 &
  21.51 &
  - &
  - &
  - &
  - \\
Pens &
  3.44 &
  \textcolor{blue}{2.75} &
  4.26 &
  11.80 &
  \textcolor{red} {2.50} &
  3.85 &
  6.49 &
  - &
  - &
  - &
  - \\
Pillows &
  0.29 &
  \textcolor{red} {0.14} &
  1.25 &
  \textcolor{blue}{0.19} &
  0.26 &
  2.26 &
  9.63 &
  - &
  - &
  - &
  - \\
Platonic &
  0.50 &
  \textcolor{red}{0.33}  &
  2.26 &
  2.78 &
  \textcolor{blue}{0.37} &
  0.82 &
  9.08 &
  - &
  - &
  - &
  - \\
Rosemary &
  \textcolor{red} {11.21} &
  \textcolor{blue}{11.75} &
  16.14 &
  26.70 &
  20.57 &
  55.81 &
  65.87 &
  - &
  - &
  - &
  - \\
Table &
  5.30 &
  \textcolor{red} {3.39} &
  4.02 &
  3.67 &
  4.94 &
  \textcolor{blue}{3.53} &
  13.62 &
  - &
  - &
  - &
  - \\
Tomb &
  0.19 &
  \textcolor{blue}{0.15} &
  1.14 &
  0.34 &
  \textcolor{red}{0.10}  &
  0.17 &
  0.44 &
  - &
  - &
  - &
  - \\
Tower &
  4.94 &
  \textcolor{red}{3.06}  &
  223.09 &
  25.28 &
  \textcolor{blue}{4.63} &
  6.35 &
  34.57 &
  - &
  - &
  - &
  - \\
Town &
  0.48 &
  \textcolor{red} {0.33} &
  1.18 &
  0.69 &
  \textcolor{blue}{0.41} &
  0.93 &
  1.20 &
  - &
  - &
  - &
  - \\
Vinyl &
  \textcolor{red} {2.97} &
  20.75 &
  20.72 &
  \textcolor{blue}{4.18} &
  7.68 &
  11.87 &
  7.05 &
  - &
  - &
  - &
  - \\
Backgammon &
  13.63 &
  \textcolor{red} {1.81}&
  6.46 &
  30.76 &
  \textcolor{blue}{3.23} &
  11.92 &
  22.05 &
  2.77 &
  1.84 &
  2.58 &
  8.81 \\
Dots &
  11.74 &
  15.64 &
  13.44 &
  \textcolor{blue} {4.89} &
  11.43 &
  \textcolor{red}{3.27} &
  182.24 &
  0.81 &
  0.88 &
  1.72 &
  21.73 \\
Pyramids &
  0.13 &
  \textcolor{blue} {0.04} &
  0.73 &
  0.17 &
  0.05 &
  \textcolor{red} {0.02} &
  0.26 &
  0.00 &
  0.00 &
  0.01 &
  0.11 \\
Stripes &
  1.45 &
  \textcolor{red} {0.74} &
  9.40 &
  3.65 &
  6.85 &
  \textcolor{blue} {0.77} &
  1043.84 &
  0.41 &
  0.21 &
  0.29 &
  6.09 \\ \midrule
Average &
  \textcolor{red} {4.34} &
  6.60 &
  25.02 &
  12.79 &
  \textcolor{blue}{5.81} &
  9.37 &
  73.83 &
  0.98 &
  0.91 &
  1.48 &
  6.01 \\ \midrule
Buddha &
  \textcolor{blue}{0.48} &
  0.49 &
  0.48 &
  1.08 &
  \textcolor{red} {0.45} &
  0.51 &
  1.25 &
  0.44 &
  0.33 &
  0.36 &
  0.32 \\
Buddha2 &
  \textcolor{red} {0.25} &
  1.32 &
  \textcolor{blue}{0.32} &
  0.66 &
  0.98 &
  3.03 &
  0.48 &
  3.56 &
  6.06 &
  6.64 &
  0.68 \\
Horses &
  \textcolor{blue}{0.62} &
  \textcolor{red}{0.57}  &
  128.05 &
  0.84 &
  1.24 &
  2.79 &
  1.97 &
  11.07 &
  6.32 &
  7.35 &
  1.60 \\
Medieval &
  \textcolor{red}{0.50}  &
  0.95 &
  26.28 &
  \textcolor{blue}{0.66} &
  0.92 &
  0.90 &
  1.44 &
  1.79 &
  1.40 &
  2.28 &
  2.05 \\
Monas &
  \textcolor{blue}{0.45} &
  0.46 &
  \textcolor{red}{0.44} &
  1.33 &
  0.56 &
  0.98 &
  7.89 &
  0.74 &
  0.79 &
  1.33 &
  0.54 \\
Papillon &
  0.69 &
  0.84 &
  \textcolor{red}{0.54}  &
  1.36 &
  \textcolor{blue}{0.67} &
  1.61 &
  11.78 &
  2.25 &
  4.98 &
  6.12 &
  1.27 \\
Stilllife &
  1.86 &
  \textcolor{red} {1.06} &
  17.07 &
  3.46 &
  \textcolor{blue}{1.38} &
  3.15 &
  12.34 &
  9.92 &
  14.07 &
  2.43 &
  2.13 \\ \midrule
Average &
  \textcolor{red}{0.69} &
  \textcolor{blue}{0.81} &
  24.74 &
  1.34 &
  0.89 &
  1.85 &
  5.31 &
  4.25 &
  4.85 &
  3.79 &
  1.23 \\ \bottomrule
\end{tabular}
\label{tab:Allexp}
\end{table*}

\begin{table*}[!t]\small
\centering
\caption{Quantitative comparisons (Q25) of different methods on HCI Blender \cite{wanner} and HCI LF benchmark dataset \cite{RN148}.  (a) Ours, (b) Han \emph{et al}. \cite{han}, (c) Williem \emph{et al}. \cite{williem},(d) Chen \emph{et al}. \cite{chen}, (e) Zhang \emph{et al}. \cite{zhang}, (f) Mishiba \emph{et al}. \cite{Mishiba}, (g) Jeon \emph{et al}. \cite{jeon}, (h) Wang \emph{et al}. \cite{wangL}, (i) Tsai \emph{et al}. \cite{tsai},(j) Shin \emph{et al}. \cite{Shin}, (k) Jin \emph{et al}. \cite{jinjing}. The best and second best results of non-learning-based methods are highlighted in \textcolor{red}{red} and \textcolor{blue}{blue}, respectively. 16 scenes (from Antinous to Vinyl) are commonly adopted as training set for learning-based methods, so we mark them as '-'.}
\begin{tabular}{@{}l|lllllll|llll@{}}
\toprule
& \multicolumn{7}{c|}{Non-learning-based}        & \multicolumn{4}{c}{Learning-based} \\
 &
  (a)&
  (b)&
  (c)&
  (d)&
  (e)&
  (f)&
  (g)&
  (h)&
  (i)&
  (j)&
  (k)\\ \midrule
Boxes &
  \textcolor{red}{0.36}  &
  0.87 &
  1.61 &
  3.28 &
  0.87 &
  \textcolor{blue}{0.86} &
  0.94 &
  0.28 &
  0.21 &
  0.44 &
  1.13 \\
Cotton &
  \textcolor{red}{0.34} &
  0.61 &
  17.08 &
  3.48 &
  0.52 &
  \textcolor{blue} {0.36} &
  0.92 &
  0.12 &
  0.10 &
  0.32 &
  0.77 \\
Dino &
   \textcolor{red}{0.53} &
  0.65 &
  2.68 &
  3.19 &
  0.63 &
  \textcolor{blue}{0.56} &
  1.03 &
  0.22 &
  0.15 &
  0.26 &
  1.14 \\
Sideboard &
  \textcolor{blue}{0.66} &
  0.99 &
  1.04 &
  3.48 &
  \textcolor{red} {0.65} &
  \textcolor{blue}{0.66} &
  0.90 &
  0.20 &
  0.16 &
  0.40 &
  1.23 \\
Antinous &
  1.12 &
  1.67 &
  13.46 &
  3.68 &
  \textcolor{blue} {1.07} &
  \textcolor{red} {0.62} &
  1.12 &
  - &
  - &
  - &
  - \\
Boardgames &
  \textcolor{red}{0.66} &
  0.76 &
  0.71 &
  1.59 &
  0.75 &
  \textcolor{blue} {0.70} &
  \textcolor{blue} {0.70} &
  - &
  - &
  - &
  - \\
Dishes&
  \textcolor{blue}{0.92} &
  1.07 &
  28.98 &
  3.00 &
  0.94 &
  \textcolor{red} {0.61} &
  1.51 &
  - &
  - &
  - &
  - \\
Greek &
  \textcolor{red} {0.55} &
  1.34 &
  26.82 &
  3.73 &
  1.32 &
  \textcolor{blue}{0.88} &
  1.77 &
  - &
  - &
  - &
  - \\
Kitchen &
  \textcolor{blue}{0.70} &
  0.96 &
  19.96 &
  1.74 &
  0.75 &
  \textcolor{red} {0.56} &
  0.96 &
  - &
  - &
  - &
  - \\
Medieval2 &
  \textcolor{red}{0.51} &
  0.58 &
  0.71 &
  1.10 &
  0.62 &
  \textcolor{blue} {0.56} &
  0.93 &
  - &
  - &
  - &
  - \\
Museum &
  0.74 &
  0.58 &
  16.30 &
  1.21 &
  \textcolor{blue} {0.51} &
  \textcolor{red}{0.46} &
  0.79 &
  - &
  - &
  - &
  - \\
Pens &
  0.56 &
  0.62 &
  \textcolor{blue} {0.43} &
  1.50 &
  0.44 &
  \textcolor{red}{0.38} &
  1.11 &
  - &
  - &
  - &
  - \\
Pillows &
  0.58 &
  \textcolor{blue} {0.56} &
  2.13 &
  1.14 &
  0.61 &
  \textcolor{red}{0.33} &
  0.63 &
  - &
  - &
  - &
  - \\
Platonic &
  0.60 &
  \textcolor{blue}{0.42}  &
  0.58 &
  1.43 &
  0.44 &
  \textcolor{red}{0.36} &
  0.46 &
  - &
  - &
  - &
  - \\
Rosemary &
  \textcolor{red} {0.74} &
  0.90 &
  1.41 &
  1.96 &
  \textcolor{blue}{0.87} &
  2.29 &
  2.68 &
  - &
  - &
  - &
  - \\
Table &
  0.88 &
  0.77 &
  1.03 &
  1.75 &
  \textcolor{blue} {0.68} &
  \textcolor{red}{0.66} &
  0.82 &
  - &
  - &
  - &
  - \\
Tomb &
  0.56 &
  \textcolor{blue}{0.55} &
  0.93 &
  1.60 &
  \textcolor{red}{0.40}  &
  0.64 &
  0.92 &
  - &
  - &
  - &
  - \\
Tower &
  \textcolor{red}{0.27} &
  0.98  &
  30.36 &
  2.89 &
  0.99 &
  \textcolor{blue}{0.66} &
  1.38 &
  - &
  - &
  - &
  - \\
Town &
  0.64 &
  \textcolor{red} {0.55} &
  0.83 &
  1.23 &
  \textcolor{red}{0.55} &
  \textcolor{blue}{0.56} &
  0.81 &
  - &
  - &
  - &
  - \\
Vinyl &
  1.03 &
  0.88 &
  8.37 &
  1.61 &
  \textcolor{blue}{0.84} &
  \textcolor{red} {0.72} &
  \textcolor{blue}{0.84} &
  - &
  - &
  - &
  - \\
Backgammon &
  1.18 &
  \textcolor{blue} {0.46}&
  1.12 &
  3.51 &
  0.50 &
  \textcolor{red}{0.34} &
  0.76 &
  0.17 &
  0.10 &
  0.17 &
  1.15 \\
Dots &
  1.17 &
  2.81 &
  40.55 &
  5.98 &
  \textcolor{red}{0.55} &
  0.97 &
  \textcolor{blue} {0.78} &
  0.24 &
  0.14 &
  0.44 &
  3.40 \\
Pyramids &
  0.88 &
  \textcolor{blue} {0.84} &
  3.12 &
  2.71 &
  1.13 &
  \textcolor{red} {0.58} &
  0.90 &
  0.13 &
  0.08 &
  0.37 &
  0.87 \\
Stripes &
  0.33 &
  \textcolor{red} {0.00} &
  \textcolor{red} {0.00} &
  2.19 &
  \textcolor{blue} {0.26} &
  0.41 &
  3.71 &
  0.17 &
  0.12 &
  0.19 &
  2.52 \\ \midrule
Average &
  \textcolor{blue} {0.69} &
  0.85 &
  9.18 &
  2.46 &
  0.70 &
  \textcolor{red}{0.66} &
  1.14 &
  0.18 &
  0.11 &
  0.29 &
  1.99 \\ \midrule
Buddha &
  0.77 &
  0.54 &
  \textcolor{red} {0.40} &
  3.51 &
  0.43 &
  \textcolor{blue}{0.42} &
  0.86 &
  0.34 &
  0.28 &
  0.38 &
  0.65 \\
Buddha2 &
  \textcolor{red} {0.68} &
  1.20 &
  \textcolor{blue}{0.70} &
  1.68 &
  1.17 &
  1.49 &
  1.19 &
  1.96 &
  1.39 &
  1.52 &
  1.78 \\
Horses &
  1.02 &
  1.11  &
  95.18 &
  2.14 &
  \textcolor{red}{0.73} &
  \textcolor{blue}{0.88} &
  1.28 &
  1.05 &
  0.92 &
  0.87 &
  1.55 \\
Medieval &
  \textcolor{blue}{0.93}  &
  1.13 &
  38.58 &
  1.86 &
  \textcolor{red}{0.75} &
  0.99 &
  1.27 &
  1.37 &
  1.44 &
  1.12 &
  1.17 \\
Monas &
  \textcolor{red}{0.30} &
  \textcolor{red}{0.30} &
  0.62 &
  3.67 &
  \textcolor{blue}{0.32} &
  0.41 &
  1.06 &
  0.49 &
  0.28 &
  0.27 &
  0.84 \\
Papillon &
  \textcolor{red}{0.87} &
  1.07 &
  \textcolor{blue}{0.93} &
  2.42 &
  1.40 &
  1.06 &
  1.27 &
  1.24 &
  1.53 &
  1.50 &
  1.65 \\
Stilllife &
  \textcolor{blue}{1.12} &
  1.23 &
  10.83 &
  3.18 &
  1.20 &
  \textcolor{red} {0.85} &
  1.40 &
  0.88 &
  0.85 &
  1.00 &
  1.33 \\ \midrule
Average &
  \textcolor{red}{0.81} &
  0.94 &
  21.03 &
  2.64 &
  \textcolor{blue}{0.86} &
  0.87 &
  1.19 &
  1.05 &
  0.95 &
  0.95 &
  1.28 \\ \bottomrule
\end{tabular}
\label{tab:Q25}
\end{table*}

\begin{table}[!t]\footnotesize 
\centering
\caption{The average MSE (100$\times$) over occluded regions (Occ) and texture-less regions (TLR)}
\begin{tabular}{@{}cll|ll|ll@{}}
\toprule
\multicolumn{3}{c|}{\multirow{2}{*}{Methods}}                                          & \multicolumn{2}{c|}{HCI \cite{RN148}} & \multicolumn{2}{c}{HCIold \cite{wanner}} \\
\multicolumn{3}{c|}{}                                  & Occ    & Tlr   & Occ   & Tlr   \\ \midrule
\multicolumn{2}{c}{\multirow{7}{*}{Non-Learning}}   & Proposed                         & 20.62       & 3.90       & 7.68         & 0.49        \\
\multicolumn{2}{c}{} & OAVC \cite{han}     & 23.53  & 22.55 & 9.62  & 0.87  \\
\multicolumn{2}{c}{} & CAE \cite{williem} & 55.13  & 20.82 & 30.90 & 29.78 \\
\multicolumn{2}{c}{} & POBR \cite{chen}    & 48.50  & 5.78  & 11.90 & 0.50  \\
\multicolumn{2}{c}{} & SPO \cite{zhang}   & 21.13  & 3.94  & 8.83  & 1.93  \\
\multicolumn{2}{c}{} & OCC \cite{wang}    & 37.26  & 8.93  & 11.98 & 3.95  \\
\multicolumn{2}{c}{} & ACC \cite{jeon}    & 125.33 & 12.06 & 20.44 & 12.34 \\ \midrule
\multicolumn{2}{c}{\multirow{4}{*}{Learning-based}} & DistgDisp\cite{wangL}    & 9.15        & 3.58       & 13.91        & 13.88       \\
\multicolumn{2}{c}{} & LFattNet \cite{tsai}    & 5.05   & 2.35  & 11.06 & 19.29 \\
\multicolumn{2}{c}{} & EPINet \cite{Shin}    & 2.49   & 2.33  & 3.64  & 4.37  \\
\multicolumn{2}{c}{} & OccUnNet \cite{jinjing} & 32.17  & 6.94  & 11.35  & 1.23  \\ \bottomrule
\end{tabular}
\label{tab:compareOccTlm}
\end{table}
\subsection{Global Depth Optimization }
\label{sec:disFinal}

Finally, we incorporate an optimization module, based on \cite{chen}, to refine the depth map in a global manner and rectify errors arising from noise, occlusions, and other factors. The optimization process employs a globally regularized scheme, minimizing an energy function defined as
\begin{equation}
\begin{split}
&E\left( d^{'} \right) = \\
&\sum_{\left( x,y \right)}^{}\begin{pmatrix}
w\left( x,y \right)\left| \left| d\left( x,y \right) - d^{'}\left( x,y \right) \right| \right|^{2} + \\
\alpha \cdot s\left( x,y \right)
\sum_{\left( x^{'},y^{'} \right)}^{}\left| \left| d^{'}\left( x,y \right) - d^{'}\left( x^{'},y^{'} \right) \right| \right|^{2} 
\end{pmatrix}.
\end{split}
\label{equ:24}
\end{equation}

The first term in Eq. (\ref{equ:24}) aims to preserve the optimized depth by mitigating distortion. To achieve this, we update the confidence in regions where significant depth gradients exist, as these gradients often contribute to depth estimation errors. In this term, \(d\) is the refined depth in Section \ref{sec:disTlm}, \(d'\) is the optimized depth, and \(w(x,y)\) is the rectified confidence, defined as 
\begin{equation}
\begin{matrix}
w\left( x,y \right) = \left\{ \begin{matrix}
c^{'}\left( x,y \right) & ,if\ std\left( d\left( x,y \right) \right) < \tau_{1}\  \\
e^{- \tau_{0}\left( \text{std}\left( d\left( x,y \right) \right) - \tau_{1} \right)} & \text{else} \\
\end{matrix} \right.
\end{matrix}   
\label{equ:20}
\end{equation}

\begin{equation}
\mathrm{c}^{\prime}(x, y)=\left\{\begin{array}{lr}
c_{o c c}(x, y), & (x, y) \in OCC \\
c_{t l}, & (x, y) \in TLR \\
c(x, y), &  otherwise
\end{array}\right.
\label{equ:25}
\end{equation}
where \(std(\cdot)\) is the standard deviation function to evaluate the depth gradient, and the standard deviation threshold \(\tau_{1}\) is set to 0.2 according to \cite{chen}. When the depth gradient is insignificant, the confidence is determined by Eq. (\ref{equ:25}) according to the point feature; otherwise, we update it by Eq. (\ref{equ:20}). \(c_{o c c}(x, y)\) is the confidence determined by Eq. (\ref{equ:12}) after half-SEPI-based refinement when the point belongs to occluded regions, \(c_{t l}\) is the constant confidence for texture-less regions and is rectified to 0.8. When the point belongs to the other regions in the LF image, its confidence is defined by Eq. (\ref{equ:12}) directly.  In addition, the amplification coefficient \(\tau_{0}\) is  set to 1000 according to \cite{chen}. 

The second term in Eq. (\ref{equ:24}) represents the smoothness term, where \(\alpha\) is the parameter that controls the trade-off between the data term and the smoothness term. The smoothness coefficient, denoted as $s(x,y)$, is defined as the product of $s_c(x,y)$ and $s_t(x,y)$, and \((x^{'},y^{'})\) represents the adjacent pixels in the 4-connected region of pixel \(\mathbf{(}x,y\mathbf{)}\). The smoothness coefficient is calculated locally for each pixel, taking into consideration its color and texture features.

\begin{enumerate}
    \item \textit{Color gradient}. 
    The smoothness of the depth estimation is closely tied to the color gradient. Regions with a low color gradient typically require smoothing of the depth due to their low confidence. Conversely, regions with high color gradients tend to preserve their depth details. The smoothness function based on the color gradient is defined as
\begin{equation}
s_{c}\left( x,y \right) = \frac{1}{\nabla I\left( x,y \right)},
\label{equ:22}
\end{equation}
where \(I\) is the central sub-aperture image.

\item \textit{Texture features:} 
We aim to maintain higher contrast not only along the edges with high confidence but also across the texture-less regions refined in Section 4.4. Therefore, the smoothness, which takes into account the texture features, is defined as
\begin{equation}
\mathrm{s}_{t}(\mathrm{x}, \mathrm{y})= \begin{cases}e^{-\tau_{2} c_{t l}} & (x, y) \in TLR \\ e^{-\tau_{2} c^{\prime}(x, y)}, & otherwise\end{cases}
\end{equation}
where the amplification coefficient \(\tau_{2}\) is set to 1.5 according to \cite{chen}.
\end{enumerate}

The final depth can be derived by solving Eq. (\ref{equ:24}) as a weighted least squares problem
\cite{RN147}.

\begin{figure*}[!t]
\centering
\includegraphics[width=18.1cm]{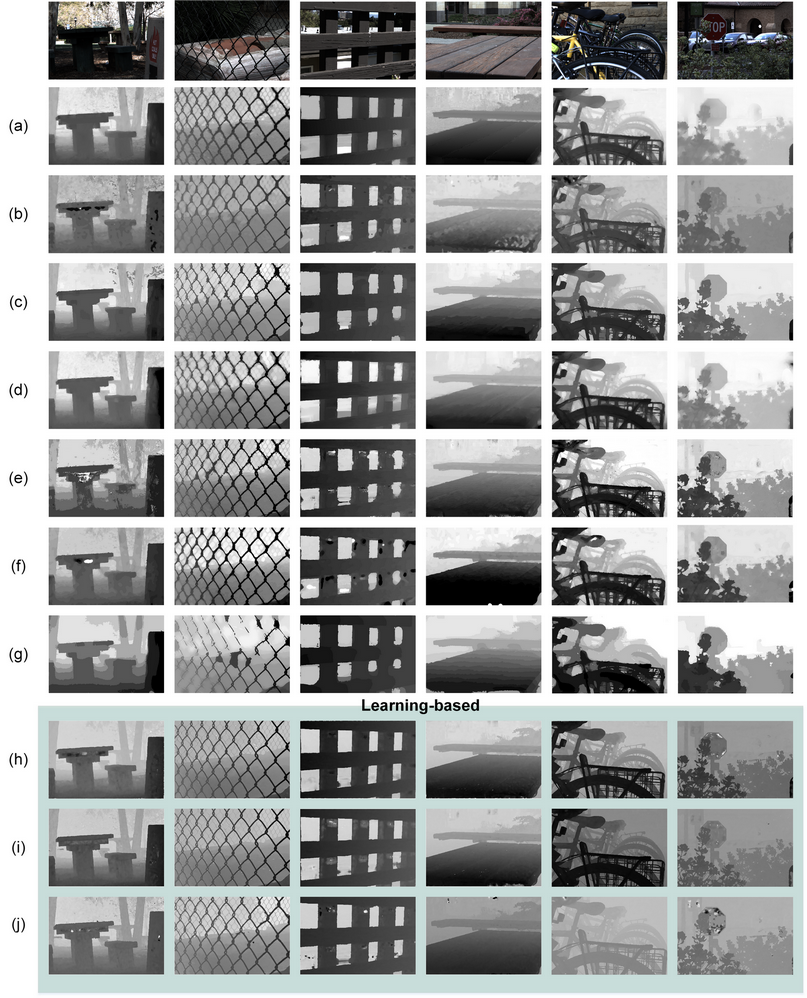}
\caption{Visual comparison of estimated depth maps on real-world LF data. (a) Ours; (b) Han \emph{et al}. \cite{han}; (c) Williem \emph{et al}. \cite{williem}; (d) Chen \emph{et al}. \cite{chen}; (e) Zhang  \emph{et al}. \emph{et al}. \cite{zhang}; (f) Mishiba \emph{et al}. \cite{Mishiba}; (g) Jeon \emph{et al}. \cite{jeon}; (h) Wang \emph{et al}. \cite{wangL}; (i) Tsai \emph{et al}. \cite{tsai}; (j) Shin \emph{et al}. \cite{Shin}.
We also refer readers to the \textit{Supplementary Material} for more results.}
\label{fig:lytro}
\end{figure*}

\begin{table*}[htbp]\small
\centering
\caption{Quantitative results of ablation studies(Overall depth map/Occluded regions/Texture-less regions).}
\centering
\renewcommand\arraystretch{1.3}
\begin{tabular}{@{}c|ccccc@{}}
\toprule

                    & \multicolumn{5}{c}{different configurations}                                            \\ \midrule
SEPI initialization 
& \(\surd\)
& \
& \(\surd\)
& \(\surd\)
& \(\surd\)  \\
Occ refinement      & \(\surd\)       & \(\surd\)       &                 & \(\surd\)       &                  \\
TLR refinement      & \(\surd\)       & \(\surd\)       & \(\surd\)       &                 &                  \\
Global optimization & \(\surd\)       & \(\surd\)       & \(\surd\)       & \(\surd\)       & \(\surd\)                 \\ \hline
HCI \cite{RN148}     & 4.34/20.62/3.90 & 9.49/46.48/4.28 & 8.11/36.15/3.97 & 4.69/33.24/3.95 & 13.70/57.83/9.48 \\
HCIold \cite{wanner} & 0.69/7.68/0.49  & 1.97/9.35/2.21  & 0.95/8.43/0.78  & 0.77/8.25/0.65  & 1.84/28.55/0.90  \\ \bottomrule
\end{tabular}
\label{tab:alaAll}
\end{table*}

\section{Experiments}
\label{sec:exp}


In our experiments, we set the number of pixels in a superpixel as $15 \times 15$ for sub-aperture images with a size of $512 \times 512$, and $17 \times 17$ for images larger than $512 \times 512$. We conducted comparisons using both synthetic and real-world LF datasets. The synthetic datasets included those provided by HCI Blenderlight field dataset \cite{wanner}  and HCI LF benchmark dataset \cite{RN148}. These datasets consist of LF images with an angular size of $9\times 9$ and corresponding ground-truth disparity maps. Additionally, we utilized real-world LF images from the Stanford Lytro LF Archive \cite{RN151}, which were captured using a hand-held LF camera, Lytro. For comparisons, we focused on the central $9\times 9$ sub-aperture images.


In our method, the adaptive threshold for identifying texture-less regions and the confidence parameter for the global optimization in such regions are crucial. To identify texture-less regions, we sorted the LTI in ascending order and determined the threshold based on the first-order difference of LTI. Since we have incorporated refinement for texture-less regions, a high and constant confidence value of 0.8 is set to prevent excessive smoothness.


\subsection{Comparisons with State-of-the-Art Methods}

We compared our method with state-of-the-art techniques, including six non-learning-based methods: Han \emph{et al}. \cite{han}, Williem \emph{et al}. \cite{williem}, Chen \emph{et al}. \cite{chen}, Zhang \emph{et al}. \cite{zhang}, Mishiba \emph{et al}. \cite{Mishiba}, and Jeon \emph{et al}. \cite{jeon}. Additionally, we compared against three supervised learning-based methods: Wang \emph{et al}. \cite{wangL}, Tsai \emph{et al}. \cite{tsai}, and Shin \emph{et al}. \cite{Shin}. Furthermore, we evaluated one unsupervised learning-based method proposed by Jin \textit{et al}. \cite{jinjing}. To ensure fair comparisons, we utilized the source codes provided by the respective authors. The parameters for the non-learning-based methods were set based on suggestions from their source codes. For training, we selected 16 samples from the HCI LF Benchmark dataset \cite{RN148}. Additionally, we constructed three test sets: one containing the remaining eight samples from the HCI LF Benchmark dataset, another consisting of seven samples from HCI Blender \cite{wanner}, and a third comprising 30 real-world LF images from the Stanford Lytro LF Archive \cite{RN151}.

\subsubsection{Quantitative comparisons}


To compare different methods quantitatively, we computed the mean square error (MSE) and the 25th percentile of the disparity errors (Q25) between the estimated disparity maps and the ground-truth ones.

As presented in Table \ref{tab:Allexp}, our method outperforms other non-learning methods in terms of depth estimation for both datasets \cite{wanner} and \cite{RN148}, with the lowest average MSE values across the entire depth map. In dataset \cite{RN148}, our method achieves an average MSE value of 4.34, which demonstrates a remarkable 25.3\% improvement compared to the second-best result, namely 5.81 from Zhang \emph{et al}. \cite{zhang}. Similarly, in dataset \cite{wanner}, our method achieves an average MSE value of 0.69, indicating a significant 14.8\% improvement over the second-best method by Han \emph{et al}. \cite{han}. Specifically, our method attains the best and second-best scores in 7 out of 24 cases, and the third-best score in 10 out of 24 cases within dataset \cite{RN148}. In dataset \cite{wanner}, our method achieves the best and second-best scores in 5 out of 7 cases. Notably, our method consistently avoids obtaining the worst MSE values in both datasets. 


In comparison to supervised learning-based methods, i.e., Wang \emph{et al}. \cite{wangL}, Tsai \emph{et al}. \cite{tsai}, and Shin \emph{et al}. \cite{Shin}, our method achieves similar but slightly inferior results in dataset \cite{RN148}, while exhibiting the best depth estimation in dataset \cite{wanner}. Additionally, when compared to the unsupervised learning-based method Jin \emph{et al}. \cite{jinjing}, our approach once again achieves the highest accuracy.

As shown in Table \ref{tab:Q25} for Q25, our method achieves the best and second-best depth estimations among non-learning methods for both datasets, respectively. When compared to supervised and unsupervised learning-based methods, our method yields similar conclusions as with MSE. Learning-based methods often achieve superior performance on data that either originates from or resembles the training set, thanks to their powerful feature mapping capabilities. However, these methods tend to underperform when faced with data that does not align with the training set's distribution, causing the effectiveness of the learned pattern models to diminish in such circumstances. In contrast, our method consistently delivers comparable performance across various datasets.



Furthermore, we conducted quantitative validation to demonstrate the advantages of our method in handling occluded regions and texture-less regions. To ensure fair comparisons, the ground truth occluded regions were derived using the method in \cite{jinjingOcc}. As illustrated in Table \ref{tab:compareOccTlm}, it is evident that our method achieves the lowest MSE values among all non-learning methods for both occluded and texture-less regions. When compared to learning-based methods, they consistently outperform our method and other non-learning methods in dataset \cite{RN148}. However, in dataset \cite{wanner}, the performance of learning-based methods is notably subpar, not only in comparison to our method but also when compared to conventional methods such as OAVC \cite{han} and SPO \cite{zhang}, particularly in texture-less regions.

\subsubsection{Visual comparisons}



In Figure \ref{fig:monas}, we present a visual comparison of depth maps estimated by various methods using synthetic LF datasets from \cite{wanner} and \cite{RN148}. A clear observation is that our method produces depth maps that exhibit consistency with the ground truth, not only in occluded regions but also in texture-less areas. The error maps provide a more precise assessment of performance. While some methods may visually appear to generate acceptable depth maps, a closer examination reveals significant deviations from the ground truth, as indicated by the yellow squares in \emph{buddha2} by Willem \cite{williem}, \emph{papillon} by Tsai \cite{tsai}, and Shin \cite{Shin} (Figure \ref{fig:monas} \textcolor{red}{(b)} to \textcolor{red}{(d)}). Comparatively, when compared to learning-based methods, our approach yields superior depth maps in dataset \cite{RN148}, particularly in \emph{platonic} and \emph{medieval}. However, in dataset \cite{wanner}, such as \emph{buddha2} and \emph{papillon}, our method performs worse (Figure \ref{fig:monas} \textcolor{red}{(e)} to \textcolor{red}{(f)}). Additionally, we conduct a visual comparison of depth results using the real-world LF dataset \cite{RN151}, where ground truth is unavailable. As depicted in Figure \ref{fig:lytro}, our method also achieves satisfactory depth maps. Although learning-based methods generally perform well, particularly in occluded regions, they exhibit limitations in texture-less areas.

Our SEPI-based method leverages the inherent characteristics of the light field itself and remains independent of specific datasets, ensuring more reliable and stable depth estimation results.

\begin{figure}[!t]
\centering
\includegraphics[width=8.8cm]{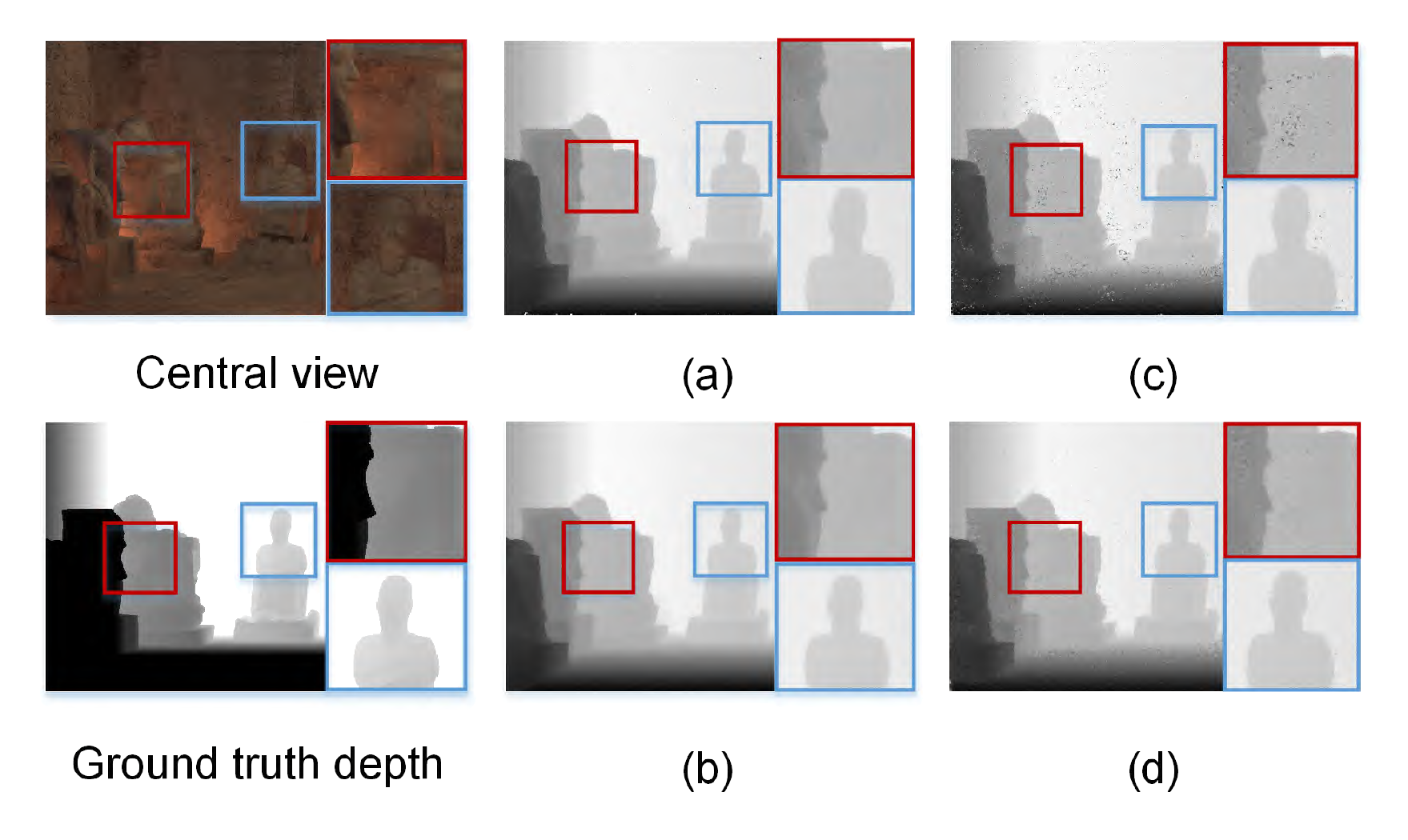}
\caption{Visual comparisons of the depth maps with/without the SEPI. (a) and (b)  are depth maps before/after global optimization with SEPI. (c) and (d) are depth maps before/after global optimization without SEPI.}
\label{fig:abla_initial}
\end{figure}

\subsection{Ablation Study}

In this section, we experimentally validated the effectiveness of the three modules contained in our framework, including the SEPI-based initial slope computation, half-SEPI-based occlusion refinement, and the refinement on texture-less regions. 
We used all LF images in  \cite{wanner} and \cite{RN148} to cover all scenarios. 

\subsubsection{Effectiveness of SEPI-based initial depth estimation}

To assess the contributions of the SEPI algorithm, we conducted an experiment where we replaced the SEPI-based initial depth estimation with the traditional EPI-based slope computation module proposed by \cite{RN61}. However, we retained the other modules, including the global optimization step. In \cite{RN61}, the initial depth estimation involves calculating the slope of the line in the EPI on slices in both horizontal and vertical directions. The result with the highest confidence among the two is then selected as the initial depth. The quantitative comparisons are illustrated in the second column of Table \ref{tab:alaAll}, and close-up images are visualized in Figure \ref{fig:abla_initial}.


When the SEPI algorithm is replaced, we observe a significant increase in MSE. This clearly demonstrates the effectiveness of the SEPI algorithm. Moreover, as shown in Figure \ref{fig:abla_initial}, the depth maps without the SEPI exhibit a substantial amount of noise. The replacement of the SEPI results in the loss of reliable initial slope information, leading to a degradation in the performance of subsequent modules such as occlusion detection, slope refinement, and depth propagation. Consequently, even the global optimization step struggles to effectively eliminate estimation errors.

\begin{figure}[!t]
\centering
\includegraphics[width=8.8cm]{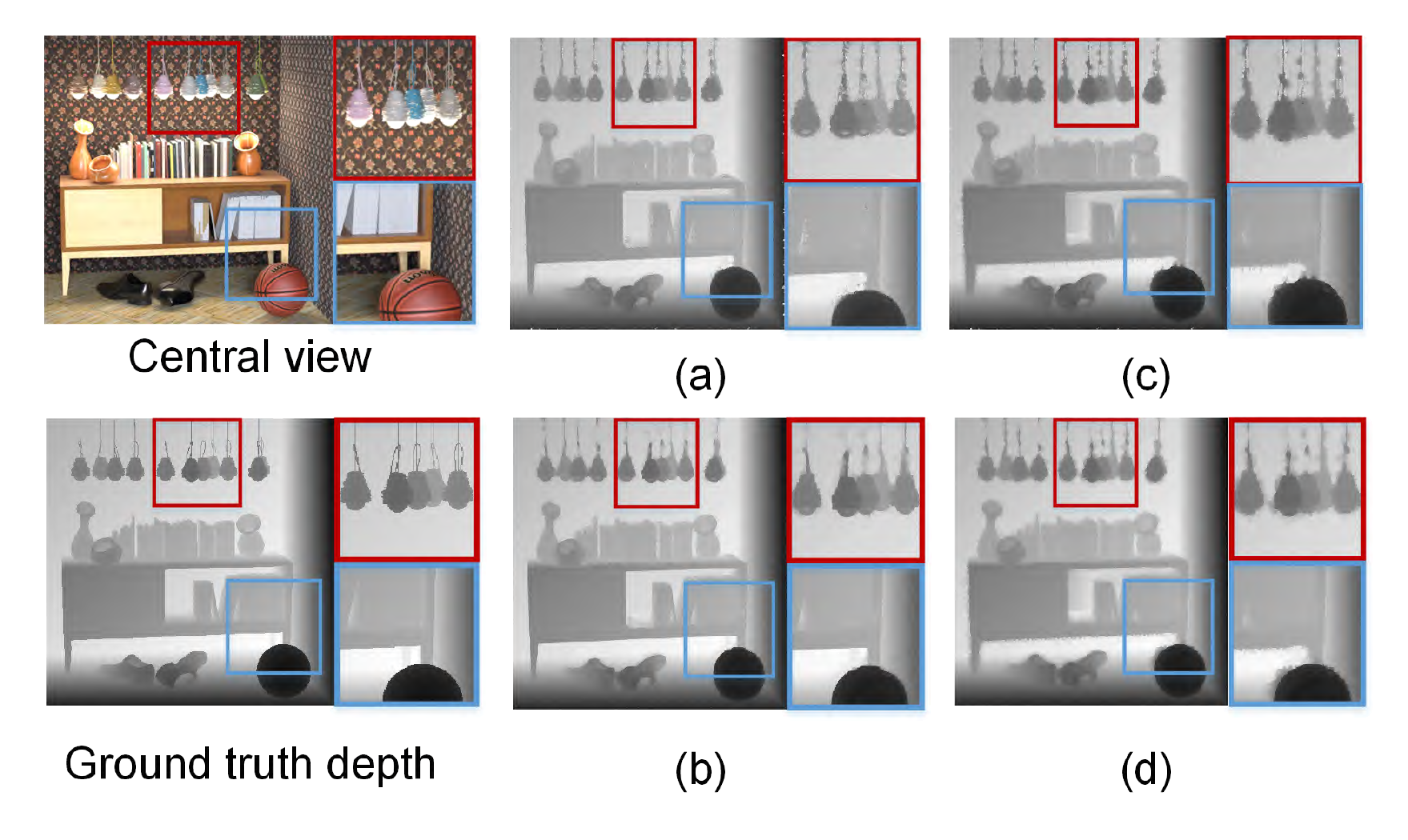}
\caption{Visual comparisons of the depth maps with/without the occlusion refinement (OR). (a) and (b)  are depth before/after global optimization with OR. (c) and (d) are depth before/after global optimization without OR. }
\label{fig:abla_occ}
\end{figure}

\subsubsection{Effectiveness of half-SEPI-based depth refinement over occluded regions}

Estimating accurate depth over occluded regions poses a significant challenge. To demonstrate the advantages of our half-SEPI-based slope refinement specifically for occluded regions, we conducted an ablation experiment where we removed the half-SEPI module and performed only the other three modules. The quantitative comparisons are presented in the third column of Table \ref{tab:alaAll}, where the MSE values exhibit a noticeable increase compared with that of the full model (the first column), indicating the effectiveness of this module. Besides, the effectiveness of this module is also validated by comparing the results shown in the fourth and fifth columns of Table \ref{tab:alaAll}, i.e., the MSE values decrease significantly when incorporating this module into the model indicated by the fifth column.


In addition, we replaced our occlusion refinement with Wang \textit{et al}. \cite{wang} to evaluate its advantage. In dataset \cite{RN148} and dataset \cite{wanner}, our method achieves average MSE values of 4.33 and 0.69, respectively, which are still superior to the corresponding values of 13.12 and 1.73 obtained by the refinement method in \cite{wang}. Moreover, since global optimization can enhance depth estimation results, we further investigated the precise benefits of the half-SEPI by presenting quantitative results specifically for occluded regions in Table \ref{tab:alaAll}. Additionally, we included intermediate close-up images in Figure \ref{fig:abla_occ}, showcasing the effects of the half-SEPI both before and after the global optimization. It is important to highlight that the removal of the half-SEPI leads to a more significant degradation in MSE for occluded regions. Furthermore, as depicted in Figures \ref{fig:abla_occ} \textcolor{red}{(c)} and \textcolor{red}{(d)}, the intermediate results without the half-SEPI exhibit more errors around occluded regions, which hinder the global optimization process in achieving higher accuracy.

\begin{figure}[!t]
\centering
\includegraphics[width=8.8cm]{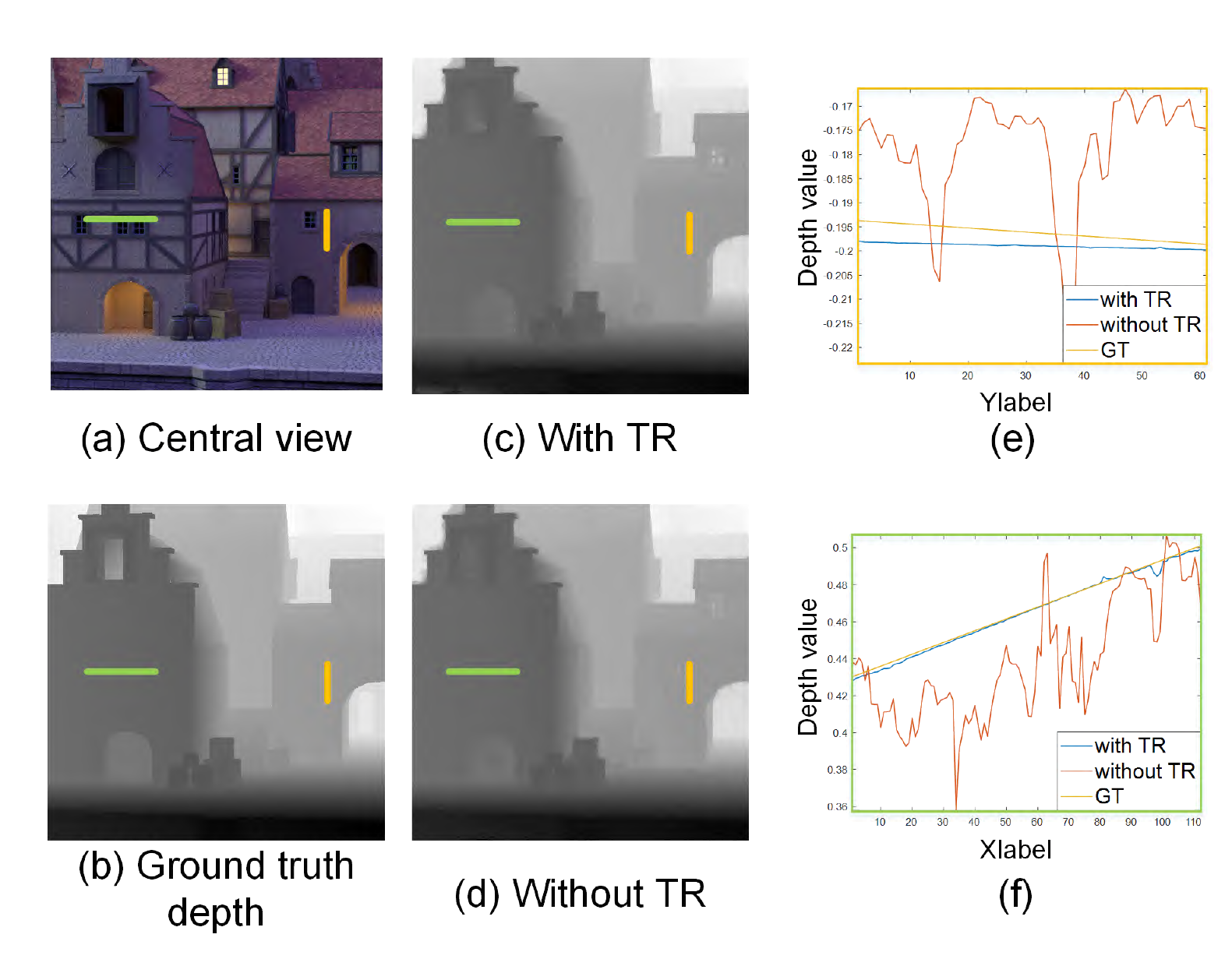}
\caption{Visual comparisons of the depth maps with/without the texture-less region refinement. (a) and (b) are depth maps with and without texture-less region refinement. (c) and (d) depth estimation results on two lines in texture-less regions with and without refinement.}
\label{fig:abla_tlr}
\end{figure}

\subsubsection{Effectiveness of depth refinement over texture-less regions}

Finally, we conducted an analysis by removing the refinement process for texture-less regions to assess its effectiveness. The quantitative comparisons are presented in the fourth column of Table \ref{tab:alaAll}. It is apparent that the MSE values experience a noticeable increase, compared with those of the full model in the first column. Besides,  the effectiveness of this module is also validated by comparing the results shown in the third and fifth columns of Table \ref{tab:alaAll}, i.e., the MSE values decrease significantly when incorporating this module into the model indicated by the fifth column. 

As depicted in Figure \ref{fig:abla_tlr}, although the depth maps obtained without refinement for texture-less regions may appear acceptable at first glance, a closer examination reveals a significant deviation from the ground truth. These comparisons vividly demonstrate the effectiveness of the refinement process for texture-less regions.

\subsection{Efficiency Analysis}

All our experiments were conducted on a PC equipped with an Intel Core i7 8700 CPU (6 cores and 12 threads), utilizing software platforms such as Matlab 2021a and MSVC 2019. We evaluated the runtime using the New HCI dataset \cite{RN148} with LF images of dimensions $512 \times 512 \times 9 \times 9$. In the absence of any parallel processing architecture,  the respective time durations for the four stages contained in our framework, i.e., initial depth estimation, half-SEPI-based refinement, refinement of texture-less regions (TLR), and global depth optimization, are approximately 1396.2s, 282.6s, 119.4s, and 2.4s. Consequently, the average total runtime is approximately 1800.6s.

Although our method currently requires a considerable amount of time, there are two important factors to consider. First, the increased time consumption results in improved performance. Second, the computations for every pixel in initial depth estimation, every region in Half-SEPI-based refinement, and TLR refinement are completely independent. Therefore, our method is well-suited for implementation with GPU-based parallel processing approaches. We intend to focus on this aspect in future work.


\section{Conclusion and discussion}
\label{sec:con}


In summary, we have introduced a novel representation for LF images called SEPI, which involves shifting and stitching EPIs corresponding to points in the 3D scene. Based on SEPI, we have proposed an LF depth estimation method that yields reliable initial depth results and improves the accuracy of depth estimation. To address occlusions, we have introduced the Half-SEPI approach, which modifies the SEPI based on occlusion properties. Furthermore, for texture-less regions with subtle depth changes, we have presented a propagation framework for depth refinement. The experimental results have demonstrated the superior performance of our method compared to state-of-the-art approaches, as evidenced by both qualitative and quantitative evaluations. Our contributions pave the way for advancing depth estimation in LF and hold promise for various applications in computer vision and related fields.

While the current implementation of our method demands a substantial amount of time, its compatibility with GPU-based parallel processing approaches makes it a promising candidate for future efficiency improvements. In our future research, we plan to explore these avenues further. Additionally, other promising directions include enhancing depth estimation along occluded boundaries using the half-SEPI approach and improving the accuracy of depth propagation over texture-less regions.




\ifCLASSOPTIONcaptionsoff
  \newpage
\fi

\bibliographystyle{IEEEtran}
\bibliography{IEEEabrv,citationlist}


\begin{IEEEbiography}[{\includegraphics[width=1in,height=1.25in,clip,keepaspectratio]{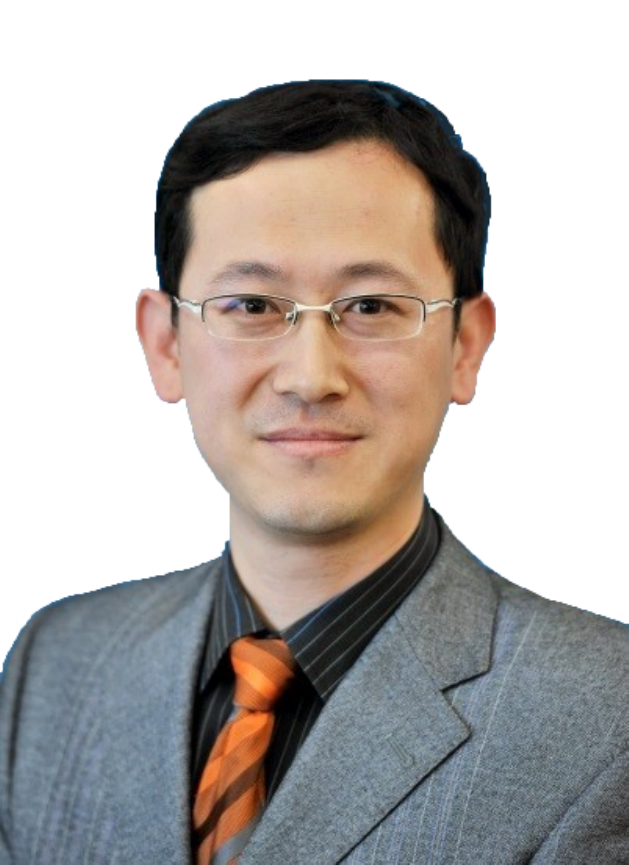}}]{Ping Zhou}
received the B.Eng. degree in Electronic Engineering from University of Science and Technology of China, Hefei, China, in 2002, and the Ph.D. degree in Biomedical Engineering from University of Science and Technology of China, Hefei, China, in 2007. He is currently an associate professor with the School of Biological Science $\&$ Medical Engineering, Southeast University. His research interests include computational imaging in biomedical engineering (light field-based system and algorithm), 3D structured-light imaging in biomedical engineering, and biomedical image processing (segmentation, classification).
\end{IEEEbiography}

\begin{IEEEbiography}[{\includegraphics[width=1in,height=1.25in,clip,keepaspectratio]{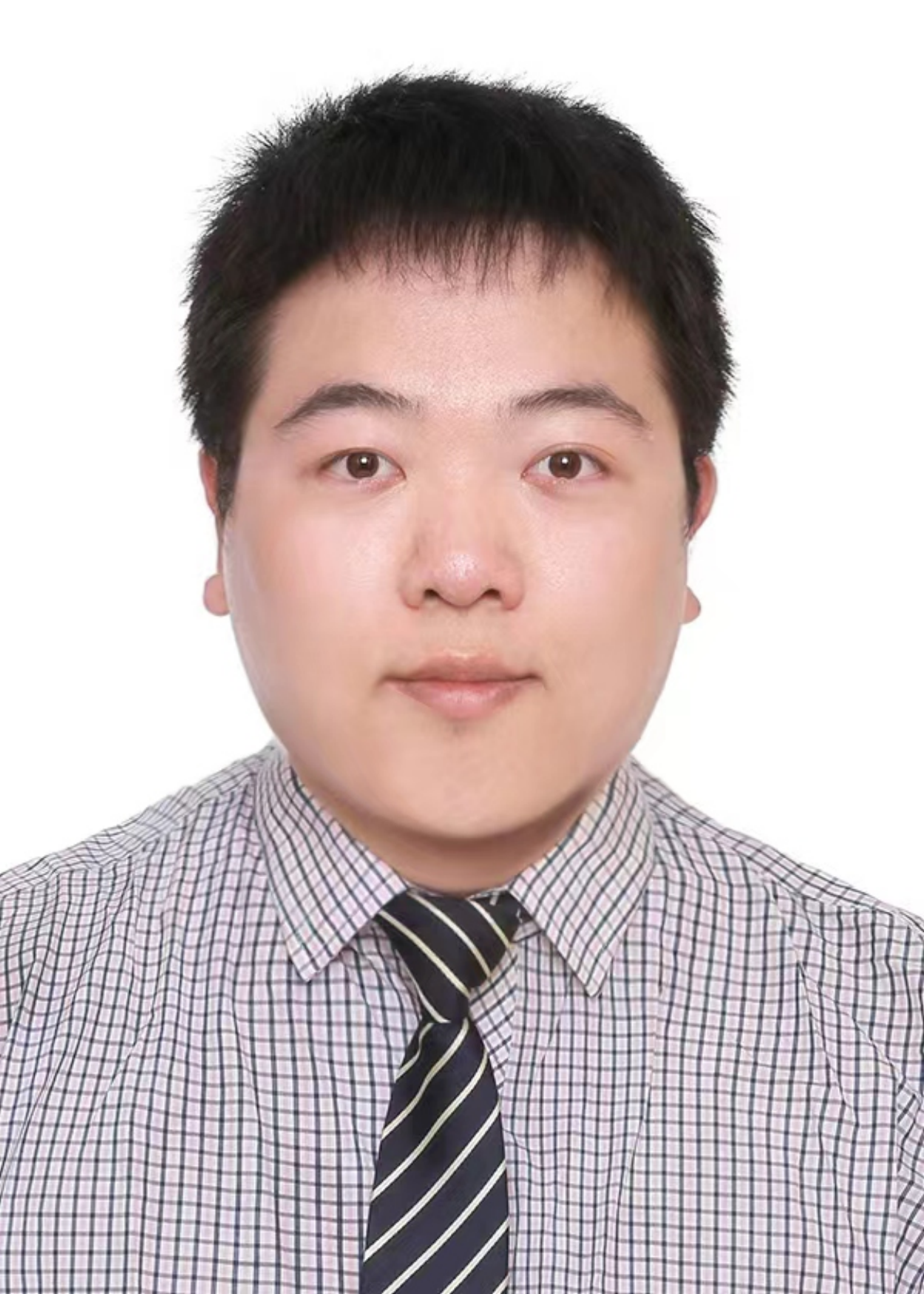}}]{Langqing Shi}
received the B.Eng. degree in Biomedical Engineering from Southeast University in 2021. He is now a postgraduate at the School of Biological Science $\&$ Medical Engineering, Southeast University. His research interests include light field image processing and super-resolution.
\end{IEEEbiography}

\begin{IEEEbiography}[{\includegraphics[width=1in,height=1.25in,clip,keepaspectratio]{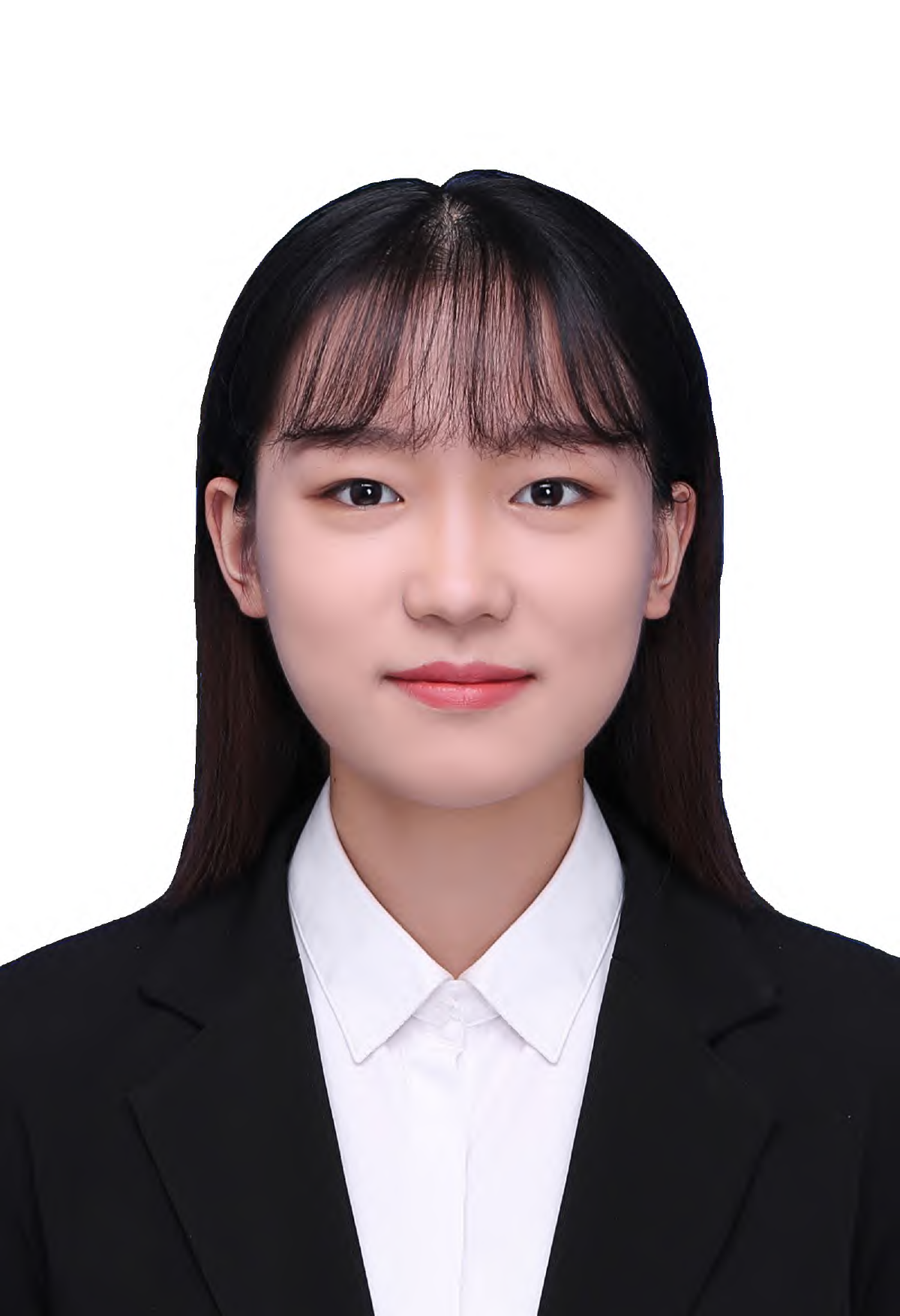}}]{Xiaoyang Liu}
received the B.Eng. degree in Biomedical Engineering from Southeast University in 2020. She is now a postgraduate at the School of Biological Science $\&$ Medical Engineering, Southeast University. Her research interests include light field image representation and processing.
\end{IEEEbiography}
\begin{IEEEbiography}[{\includegraphics[width=1in,height=1.25in,clip,keepaspectratio]{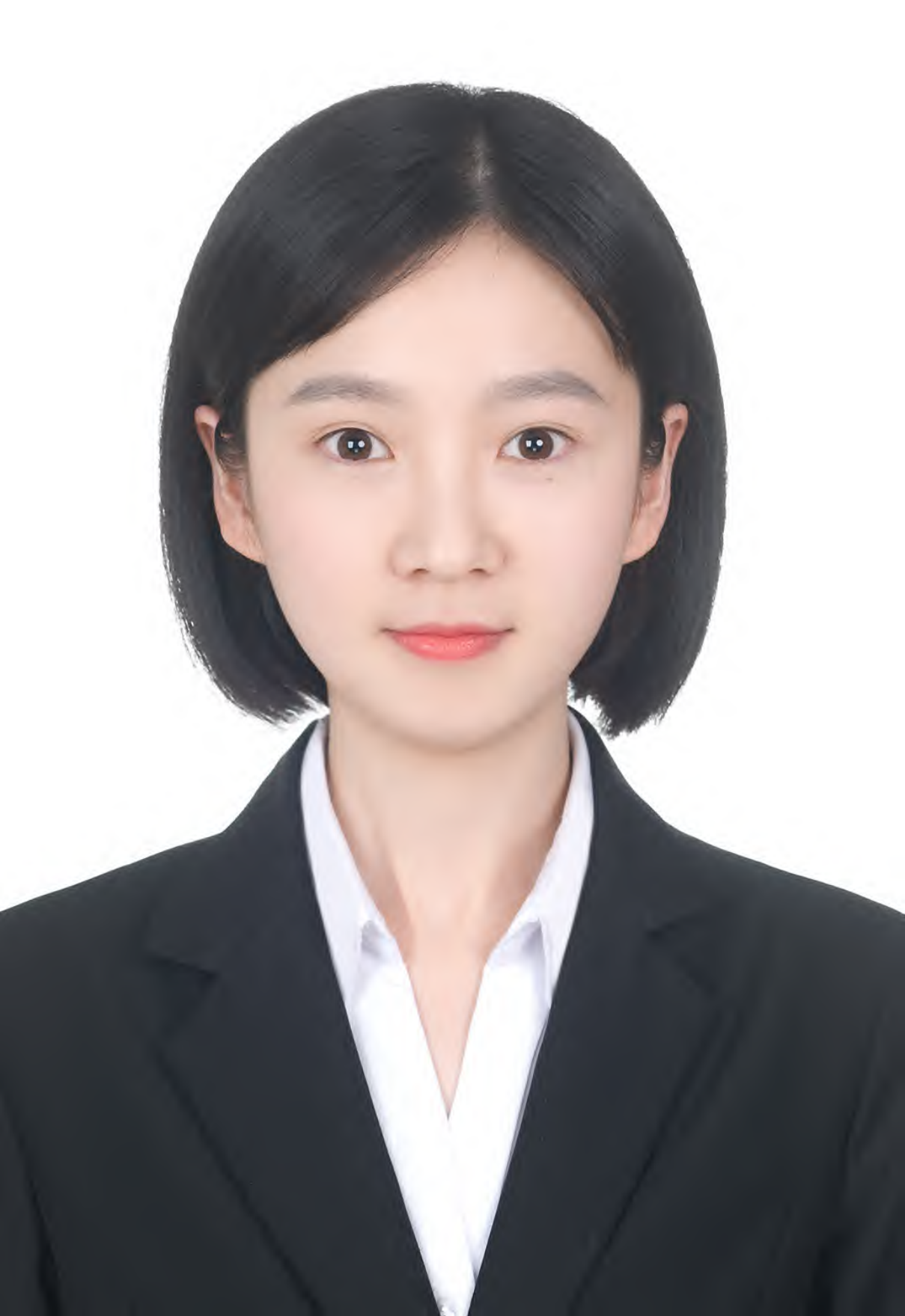}}]{Jing Jin}received the B.Eng. degree from the
Southeast University, Nanjing, China, in 2017.
She is currently pursuing the Ph.D. degree in
computer science with the City University of
Hong Kong, Hong Kong SAR. Her research interests
include light field image representation
and processing.
\end{IEEEbiography}

\begin{IEEEbiography}[{\includegraphics[width=1in,height=1.25in,clip,keepaspectratio]{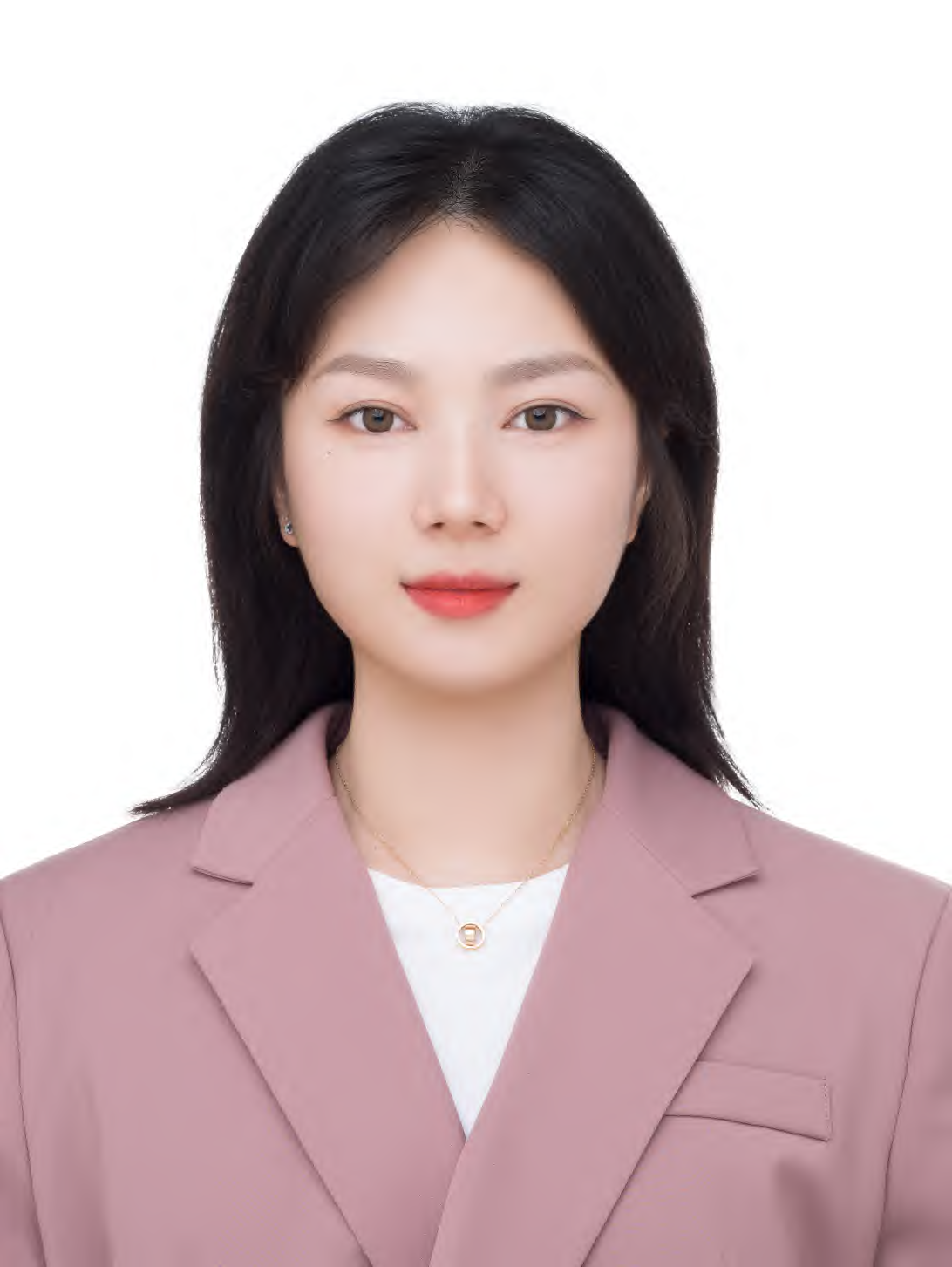}}]{Yuting Zhang}
received the B.Eng. and MA.Eng degrees in Biomedical Engineering from Southeast University in 2018 and 2021, respectively. Her research interests include 3D reconstruction, light field image representation and processing.
\end{IEEEbiography}

\begin{IEEEbiography}[{\includegraphics[width=1in,height=1.25in,clip,keepaspectratio]{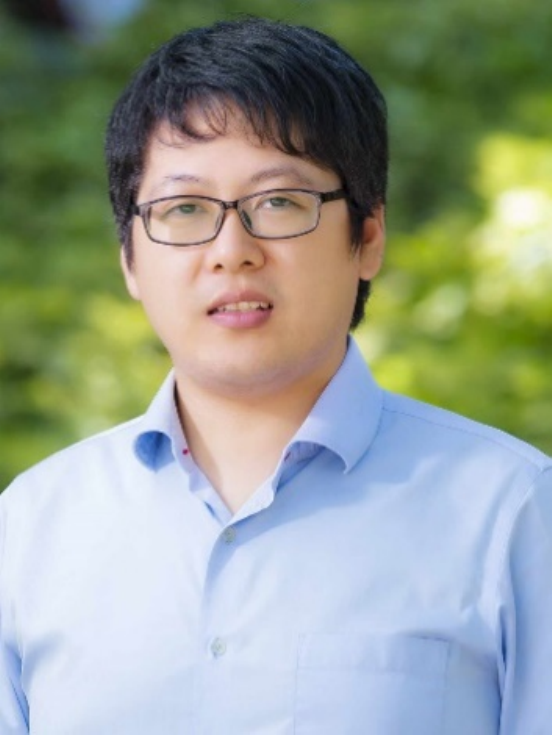}}]{Junhui Hou}
is an Associate Professor with the Department of Computer Science, City University of Hong Kong.  He holds a B.Eng. degree in information engineering (Talented Students Program) from the South China University of Technology, Guangzhou, China (2009), an M.Eng. degree in signal and information processing from Northwestern Polytechnical University, Xi’an, China (2012), and a Ph.D. degree from the School of Electrical and Electronic Engineering, Nanyang Technological University, Singapore (2016). His research interests are multi-dimensional visual computing.  
	
Dr. Hou received the Early Career Award (3/381) from the Hong Kong Research Grants Council in 2018. He is an elected member of IEEE MSA-TC, VSPC-TC, and MMSP-TC. He is currently serving as an Associate Editor for \textit{IEEE Transactions on Visualization and Computer Graphics}, \textit{IEEE Transactions on Circuits and Systems for Video Technology}, \textit{IEEE Transactions on Image Processing}, \textit{Signal Processing: Image Communication}, and \textit{The Visual Computer}.

\end{IEEEbiography}

\end{document}